\documentclass[twoside,11pt]{article}

\usepackage{blindtext}

\usepackage[preprint]{jmlr2e}

\usepackage{lastpage}
\jmlrheading{23}{2026}{1-\pageref{LastPage}}{1/22; Revised 5/22}{9/22}{21-0000}{Daniele Nikzad and Mikko Koivisto}

\ShortHeadings{Generalizing Metropolis--Hastings for DAG Sampling}{Nikzad and Koivisto}
\firstpageno{1}

\usepackage[american]{babel}
\usepackage{mathtools} % amsmath with fixes and additions
\usepackage{booktabs} % commands to create good-looking tables
\usepackage{tikz} % nice language for creating drawings and diagrams

\usetikzlibrary{arrows.meta}
\usepackage{algorithm}
\usepackage{algpseudocode}
\usepackage{amsmath}
\usepackage{graphicx}
\usepackage{amssymb}
\usepackage{float}
\usepackage{scalefnt}
\usepackage{xstring}
\usepackage{array}
\usepackage{enumitem}
\usepackage[detect-weight]{siunitx}
\makeatletter

\algrenewcommand\alglinenumber[1]{\small #1:}
\algtext*{EndFor}
\algtext*{EndIf}

\newcommand{\be}{\begin{eqnarray}}
\newcommand{\ee}{\end{eqnarray}}
\newcommand{\bes}{\begin{eqnarray*}}
\newcommand{\ees}{\end{eqnarray*}}

\newcommand{\cP}{\mathcal{P}}
\newcommand{\cQ}{\mathcal{Q}}

\newcommand{\dd}{\mathrm{d}}
\newcommand{\perms}[1]{\mathfrak{S}_{#1}}

\newcommand{\BigCapsBase}[1]{%
  {\scalefont{1.0}\MakeUppercase{\@car#1\@nil}}%
  \textsc{\MakeLowercase{\@cdr#1\@nil}}%
}

\newcommand{\BigCaps}[1]{%
  \IfBeginWith{#1}{BG}{%
    {\scalefont{1.0}\textsc{BG}}% acronym full-size
    \StrGobbleLeft{#1}{2}[\rest]%
    \textsc{\MakeLowercase{\rest}}% suffix small
  }{%
    \IfBeginWith{#1}{BMH}{%
      {\scalefont{1.0}\textsc{BMH}}% acronym full-size
      \StrGobbleLeft{#1}{3}[\rest]%
      \textsc{\MakeLowercase{\rest}}% suffix small
    }{%
      \BigCapsBase{#1}% default behavior
    }%
  }%
}
\makeatother

\newcommand{\CC}{C\nolinebreak\hspace{-.05em}\raisebox{.4ex}{\tiny\bf +}\nolinebreak\hspace{-.10em}\raisebox{.4ex}{\tiny\bf +}}
\def\CC{{C\nolinebreak[4]\hspace{-.05em}\raisebox{.4ex}{\tiny\bf ++}}}

\newcommand{\asia}{{\sc Asia}}

\newcommand{\sachs}{{\sc Sachs}}

\newcommand{\hailfinder}{{\sc Hailfinder}}

\newcommand{\andes}{{\sc Andes}}

\newcommand{\pigs}{{\sc Pigs}}

\long\def\comment#1{}

\begin{document}

\title{Generalizing Metropolis–Hastings for DAG Sampling}

\author{\name Daniele Nikzad \email daniele.nikzad@helsinki.fi \\
       \addr Department of Computer Science\\
       University of Helsinki\\
       Helsinki, FI-00014, Finland
       \AND
       \name Mikko Koivisto  \email mikko.koivisto@helsinki.fi  \\
       \addr Department of Computer Science\\
       University of Helsinki\\
       Helsinki, FI-00014, Finland}

\editor{My editor}

\maketitle

\begin{abstract}%   <- trailing '%' for backward compatibility of .sty file
A fundamental challenge in Bayesian network structure learning lies in efficiently approximating the posterior distribution of directed acyclic graphs (DAGs). For large, high-dimensional datasets, the task remains beyond the capabilities of current methods. This work advances Markov chain Monte Carlo algorithms by introducing novel moves that simultaneously resample the parents for multiple nodes of the DAG. At the core of these moves is \emph{duet sampling}, a new generalization of the Metropolis--Hastings algorithm that selects two proposal distributions from a pool at each iteration. One is used for generating a candidate state of the Markov chain and another for evaluating the return to the current state; both are ingredients of an appropriately crafted acceptance criterion. Pairing different proposal distributions provides additional freedom and enables constructing Markov chains that mix more efficiently than their basic Metropolis--Hastings counterparts, at no additional computational cost. Experiments show that our new moves, both individually and in combination, outperform the previous state of the art, which relies on a more restrictive pool of moves. This advantage persists even when letting the previous methods leverage duet sampling to enhance their performance.
\end{abstract}


\section{Introduction}\label{sec:intro}

Given the pivotal role of directed acyclic graphs (DAGs) in capturing conditional independencies \citep{Pearl1988,Koller2009} and causal structures \citep{Pearl2009},
substantial research effort has been devoted to learning DAG models from data. Since the seminal studies by \citet{Buntine1991}, \citet{Cooper1992}, and \citet{Madigan1994}, the Bayesian paradigm for statistical inference has become one of the most extensively explored approaches to learning. The Bayesian approach is particularly appealing because it provides a principled formalism for expressing uncertainties about any involved unknowns both before and after seeing the data, by prior and posterior probabilities, respectively \citep{Heckerman1995}.

Computationally, constructing DAGs that accurately represent the posterior distribution has proven to be very demanding. Even finding a single \emph{posterior mode} is NP-hard \citep{Chickering1996}, the known exact algorithms being feasible only up to around 30 nodes \citep{Koivisto2004,Silander06,Yuan2013} or, in some favorable cases, up to around a hundred of nodes \citep{Bartlett2017}. To tackle larger networks, various search heuristics have been explored, without optimality guarantees \citep{Kitson2023}. As constructing the \emph{full DAG posterior} is feasible only for DAGs on five or fewer nodes,  moderately exponential exact algorithms have been developed to compute \emph{low-dimensional marginals}, such as edge and ancestor relation probabilities  \citep{Koivisto2004,Koivisto2006,Pensar2020,Tian2009}, as well as to \emph{sampling} DAGs from the posterior \citep{Harviainen2024,He2016,Talvitie2019uai}. These algorithms are practical up to around 25 nodes. For Bayesian learning of larger DAGs, research has focused on two computational paradigms: Markov chain Monte Carlo (MCMC) methods, which are guaranteed to yield accurate approximations if run long enough \citep[e.g.,][]{Friedman2003,Goudie2016,Grzegorczyk2008,Niinimaki2016,nikzad2025,Su2016,Viinikka2020}, and variational methods that aim to find an optimal proxy posterior in some simplified class of distributions \citep{Annadani2023,Cundy2021,Lorch2021}.

Recent work \citep{nikzad2025} suggests that the most promising approach to approximate the posterior of larger networks is MCMC on the natural DAG space. Specifically, the work gives highly efficient implementations both for basic single-edge moves \citep{Madigan1995,Giudici2003} and for advanced moves that simultaneously resample the parents of two adjacent nodes \citep{Grzegorczyk2008} or of a node and its children \citep{Su2016}. For the latter move types, the key new discovery is a pruning technique that enables discarding possible parent sets as preprocessing, often significantly reducing the computations per parent set update during the Markov chain simulation. The presented algorithm, Gibby, was shown to outperform other approaches: MCMC on a node partition space \citep[BiDAG,][]{Kuipers2017,Suter2023}, variational inference in a ``continuous DAG space'' \citep[DiBS,][]{Lorch2021}, and generative flow networks \citep[DAG-GFlowNet,][]{Deleu2022}. However, Gibby also inherits a major weakness of the implemented moves: even when resampling entire parent sets, the moves are constrained to preserve certain adjacencies in the local structure. Another promising option is the \emph{blocked Gibbs} (BG) move of \cite{Goudie2016}, which resamples the parents for a few randomly selected nodes from their joint full conditional posterior, making so effective use of the local posterior and yielding an efficiently mixing Markov chain. The computational cost of the BG move can likewise be reduced by leveraging the pruning technique of \cite{nikzad2025}. Nevertheless, the computations quickly become prohibitively expensive, unless the block size and the number of parents per node are bounded by small constants. 

Motivated by the limitations of existing moves, we ask: \emph{Are there other types of moves that enable even more flexible jumps to high-posterior DAGs without compromising computational efficiency?}

We will answer this question in the affirmative by introducing novel moves that simultaneously update the parent sets of multiple nodes. Our moves are motivated by two lines of previous work and thus can be divided into two broad classes. First, we study a move that resamples the parents for a random pair of nodes, such that the edge between the nodes is added, removed, or reversed, depending on their current adjacency. This \emph{edge-constrained}~(EC) move extends both the single-edge moves by \cite{Giudici2003} and the edge reversal move by \cite{Grzegorczyk2008}. We also present a variant, the \emph{ancestor-constrained} (AC) move, which imposes constraints on the ancestor relation rather than direct adjacency between the two nodes. The second class of moves is inspired by the BG move. Like BG, our \emph{unconstrained sequential resampling} (USR) move samples blocks of parent sets for multiple nodes without imposing any constraints other than acyclicity. However, it generates the parent sets sequentially rather than jointly, which significantly reduces the computational cost. In return, a USR move may reject the proposal. We mitigate this shortcoming by leveraging a novel MCMC algorithm that can favor long jumps in the DAG space.

Indeed, the success of the new USR move relies on a generalization of the Metropolis--Hastings (MH) algorithm that, we believe, is of independent interest. The standard MH algorithm \citep{Hastings1970,Tierney1994} uses an acceptance ratio that applies the \emph{same} proposal distribution in two directions: from the current state to the proposed state, and vice versa. Our generalization, which we call \emph{duet sampling}, employs \emph{two} proposal distributions: one in the forward direction and another in the reverse direction. By using a suitable acceptance criterion, we obtain a reversible Markov chain. The added flexibility allows us to choose pairings of proposal distributions that enhance the mixing properties of the Markov chain. This turns out to be crucial for the efficiency of the USR move, which would perform substantially worse if implemented under the standard Metropolis--Hastings algorithm; we also observe duet sampling to yield a notable gain in (our implementation of) the Markov blanket resampling move of \citet{Su2016}. Also the EC and AC moves are naturally written in terms of duet sampling.   

The rest of this article is organized as follows. Section~\ref{sec:preliminaries} introduces the Bayesian approach to learning Bayesian networks and reviews previous algorithms based on simulating Markov chains over the space of DAGs. Section~\ref{sec:newacc} gives the formulation of duet sampling, along with two particular ways to pair proposal distributions: independently at random and deterministically. Section~\ref{sec:nm} presents the new moves, AC, EC, and USR, with concrete instantiations of the two classes of pairings.  Section~\ref{sec:exp} is dedicated to empirical studies: focusing primarily on large, high-dimensional datasets, we highlight the limitations of previous algorithms and the consistent superiority of our new algorithms. Section~\ref{sec:cr} concludes and points to directions for future research. We have relegated to the appendix routine proofs and additional analyses that, while likely of secondary interest, are essential for completeness.

\section{Preliminaries}\label{sec:preliminaries}

In this section, we describe the learning problem and present the core Markov chain Monte Carlo methods, including the most recent advancements, on top of which we develop our novel algorithms. 

\subsection{Learning Bayesian Networks}

A Bayesian network (BN) represents an $n$-variate probability distribution by a pair $(G, \theta)$, where $G$ is a directed acyclic graph (DAG) encoding conditional independences (CIs) among the variables and $\theta$ specifies a distribution over the $n$ variables  compatible with the CIs. In this model, we let~$N := \{1,2,\dots,n\}$ denote the set of nodes of the graph $G$. If $ij$ is an edge in $G$, we call $i$ the \emph{parent} of $j$ and $j$ the child of $i$. We denote the sets of parents and children of $i$ by $P_i(G)$ and $C_i(G)$ respectively. The set of descendants of $i$, denoted by $D_i(G)$, consists of nodes to which there is a directed path from $i$ in $G$, including node $i$ itself. If there is no ambiguity about the graph $G$, we may write simply $P_i$, $C_i$, and~$D_i$.

To learn a Bayesian network from data over the $n$ variables, we model the data points as independent draws, yielding the likelihood of $(G, \theta)$. Taking a Bayesian approach, we also assign a prior of $(G, \theta)$. Supposing the interest is in the DAG $G$, the result of learning is the posterior distribution of $G$, obtained by the Bayes rule and by marginalizing $\theta$ out. Under commonly assumed structural properties of the prior \citep[Sec.~18.3.6]{Koller2009}, the posterior probability of~$G$, denoted by $\pi(G)$, is \emph{modular}, that is,
\be \label{eq:modular}
	\pi(G) \propto \prod_{i \in N} \pi_i(P_i)\,,
\ee
where each $\pi_i(P_i)$, which we call a \emph{local score}, is a product of a prior component and a likelihood component. While in principle the parent set $P_i$ could be any subset of $N\setminus\{i\}$, in practice we restrict $P_i$ to some family of \emph{possible parent sets} $\cP_i$, e.g., by setting some moderate upper bound for the number of parents, say, four or five. 

Common choices for the prior term include the uniform prior, which assigns equal probability to all parent sets, and sparsity-inducing priors, which assign lower probabilities as the size of the parent set increases. Closed-form expressions for the marginal likelihood are obtained by adopting likelihood models and priors that permit analytic integration. Among these, the Bayesian Dirichlet equivalent uniform (BDeu) score for discrete data \citep{Buntine1991,Heckerman1995} and the Bayesian Gaussian equivalent for continuous data (BGe) score \citep{Geiger2002,Kuipers2014} allow efficient evaluation of $\pi_i(J)$ for any given set $J$. Moreover, both BDeu and BGe are \emph{score equivalent}, assigning identical scores to DAGs that represent the same conditional independencies.

\subsection{Structure MCMC}\label{sec:MCMC}

\emph{Structure MCMC} refers to MCMC methods that are based on approximately sampling DAGs from the posterior $\pi$ by simulating an ergodic Markov chain $G^1, G^2, \ldots$ whose stationary distribution is $\pi$. For ergodicity, it suffices that the transition kernel $p_m(G^{t+1} | G^t)$ of the chain satisfies the detailed balance condition,  
\be \label{eq:db}
	\pi(G) p_m(G' | G) = \pi(G') p_m(G | G')
	\quad \textrm{for all $G \neq G'$}\,,
\ee
and that the chain is irreducible and aperiodic. Here $m$ stands for a \emph{move} that specifies the way $G$ is transformed to $G'$, e.g., by local changes of individual edges or entire parent sets. Aperiodicity is easily ensured for moves that stay in the same state with positive probability. For irreducibility, it suffices to show that any graph $G$ can be obtained from the empty graph by a sequence of steps, and vice versa.

\sloppy
A common way to guarantee detailed balance is to construct the chain using the Metropolis--Hastings (MH) algorithm \citep{Hastings1970}. Given the current state $G$, the algorithm first draws a candidate state $G'$ from a proposal distribution $q_m(G' | G)$. Then the candidate is accepted as the new state with probability  
\be \label{eq:MH}
	\alpha_m(G' | G) := \min\bigg\{1,\frac{\pi(G')q_m(G | G')}{\pi(G)q_m(G' | G)}\bigg\}\,; 
\ee
otherwise the Markov chain stays at the state $G$. When $G\neq G'$, a direct calculation shows that~$p_m(G' | G) = q_m(G' | G) \alpha_m(G' | G)$ satisfies the detailed balance condition \eqref{eq:db}. 

It is often useful to apply multiple moves $m$, each with some probability $\beta_m$ \citep{Grzegorczyk2008,nikzad2025,Su2016}. This results in a mixture proposal distribution $q(G'|G) = \sum_m \beta_m q_m(G'|G)$. However, it may be computationally demanding to evaluate the probability $q(G'|G)$, especially if the number of mixture components is large. Fortunately, one can avoid that evaluation by using the rule \eqref{eq:MH} as is, separately for each $m$. Namely,  detailed balance holds for the resulting transition kernel, since it holds separately for each move. For further discussion on these two variants, we refer to \citet[p.~63]{Besag1995} and \citet[Sec.~4]{Tierney1998}.   

In the framework reviewed above, a \emph{move} is specified by a proposal distribution $q_m$ and a \emph{mixture move} by a mixture proposal distribution $\sum_m \beta_m q_m$. We will find this viewpoint too restrictive when we generalize the MH acceptance rule in Section~\ref{sec:newacc}. Namely, our \emph{duet sampling} employs pairs of distributions, $q_k$ and $q_l$, and a mixture move gets specified only once the pairwise selection probabilities, $\beta_{kl}$, are fixed.

\subsection{Previous Move Types}\label{sec:pmt}

The moves considered in the sequel can be viewed as instantiations of the following template. A move begins by \emph{orphaning} a subset of nodes $S \subseteq N$, i.e., by removing all incoming edges to nodes in $S$ from the current DAG $G$, thereby yielding a DAG we denote by $G_0$. The choice of $S$ depends on the move under consideration. Next, a permutation $\sigma = (\sigma(1), \sigma(2), \dots, \sigma(|S|))$ of the nodes in $S$ is selected either randomly or deterministically, again depending on the move. Then, a sequence of intermediate DAGs is constructed by sequentially sampling new parent sets $P'_i$ of the orphaned nodes $i \in S$, in the order specified by $\sigma$, with probabilities proportional to their local scores $\pi_i(P'_i)$, and cumulatively adding the corresponding edges to $G_0$. The possible parent sets are restricted by move-specific constraints and the requirement to keep the intermediate DAGs acyclic. 

Thus, the proposal distribution, composed using the chain rule of probability, becomes
\begin{equation}\label{eq:q0}
	q_m^\sigma (G'|G)= \prod_{s=1}^{|S|} \frac{\pi_{\sigma(s)}(P'_{\sigma(s)})}{ Z^m_{\sigma(s)}(G'_{\sigma, s-1}) }\,, \quad \text{where} \quad
	Z_{\sigma(s)}^m(G'_{\sigma,s-1}) = \sum_{J \in \mathcal{P}_{\sigma(s)}^m(G'_{\sigma,s-1})} \pi_{\sigma(s)}(J) \,.
\end{equation}
Here, $G'_{\sigma,s-1}$ denotes the DAG obtained by incorporating the new parent sets of the first~$s-1$ nodes in~$\sigma$ into the orphaned DAG $G_0$, and $\mathcal{P}^m_{\sigma(s)}(G'_{\sigma,s-1}) \subseteq \mathcal{P}_{\sigma(s)}$ is the family of possible parent sets of node~$\sigma(s)$ that satisfy the move-specific constraints and do not introduce cycles when incorporated into $G'_{\sigma,s-1}$. When the indices $m$ or $\sigma$ can be inferred from the context, we may omit them from the notation.

The orphaning phase is crucial, even if it may seem unnecessary. Removing edges reduces the number of descendants of nodes $i \in S$, which allows certain edges or ancestor relationships to be reversed in the proposed DAG $G'$, thereby improving the mixing of the Markov chain. Without orphaning, these moves could not reverse ancestor relationships in a single iteration, potentially causing the chain to become trapped at a local mode.

Two move classes, representing the current state of the art, are instantiations of the above template. One is the \emph{edge reversal} (REV) move of \cite{Grzegorczyk2008}. It proposes the DAG $G'$  by first drawing an edge $ij$ in the current DAG $G$ uniformly at random. Then the parent sets of the two end nodes of the edge are resampled according to their score, imposing the presence of the reversed edge $ji$. The sampling order is unspecified in the original article, and the order does not matter: orders $(i,j)$ and $(j,i)$ produce the same result. However, forcing the reversal of an edge $ij$ is not always desirable, as one high-probability DAGs may contain an edge between $i$ and $j$, while another high-probability DAG does not. In this case, the REV move does not provide a straightforward transition between these DAGs, potentially leading to poor exploration of the Markov chain.

Another instantiation is the \emph{Markov blanket resampling} (MBR) move by \cite{Su2016}. It first selects a node~$i$ uniformly at random and puts $S = \{i\} \cup C_i(G)$. The nodes in this set are then orphaned, and their new parent sets are sequentially sampled according to their scores, following an order given by a randomly selected permutation $\sigma$ of $S$. The new parent sets are constrained so that node $i$ remains a parent of every node in $C_i(G)$ in the proposed DAG $G'$. However, this latter condition preserves adjacencies within $S$ and precludes edge reversals in the proposed DAG, which might hinder jumps to high probability regions of the DAG posterior.

The REV and MBR moves fall within the standard MH framework, each proposal being accepted with probability~\eqref{eq:MH}. MBR can be generalized within our duet sampling framework to enhance the mixing of the Markov chain; the details are in Appendices~\ref{sec:comp} and~\ref{sec:small},

The blocked Gibbs (BG) move of \cite{Goudie2016} does not belong to the class of moves described above. Namely, it proposes a DAG $G'$ by jointly resampling the parent sets for a block of nodes $S \subseteq N$ from the full conditional posterior, without specifying any sampling order $\sigma$. Specifically, a new parent set tuple $P'_S = (P_i')_{i \in S}$ is drawn from the conditional distribution given the parent sets of the nodes outside $S$. As a result, the proposed graph $G'$ is always accepted. Exact sampling is implemented in two phases: First, an \emph{ancestry DAG} $H$ on $S$, representing constraints on ancestor relationships and corresponding to a set of tuples $P'_S$, is sampled according to its respective (marginal) posterior distribution. Second, conditionally on $H$, each $P'_i$, for $i \in S$, is sampled independently subject to the constraints. Since the algorithm evaluates the unnormalized posterior probability of every possible DAG $H$ on $S$, whose number grows superexponentially with the size of the block, this exhaustive evaluation is feasible only for small blocks, typically of size at most three or four.

\subsection{Score Pruning and Approximating Score Sums}\label{sec:pru}

The normalizing constant $Z_{\sigma(s)}^m(G'_{\sigma,s-1})$ in \eqref{eq:q0} represents the main computational bottleneck of the class of moves described above.  It involves summing the scores over all parent sets that satisfy the move-specific constraints. In the worst case, the computational cost scales linearly in the number of possible parent sets $|\mathcal{P}_{\sigma(s)}|$ of the node. Likewise, for the BG move, the evaluation of the full conditional posterior over all blocks of possible parent sets requires the computation of a similar normalizing constant. 

To reduce the possible parent sets that need to be stored in memory and considered when computing the normalizing constants, \cite{nikzad2025} introduce a pruning technique that discards low-scoring parent sets as preprocessing, thereby significantly reducing the number of summands in the normalizing constants. The key idea is to discard a parent set if its score is lower than an appropriately weighted sum of the scores of its subsets, scaled by an arbitrarily chosen parameter $\epsilon \in [0, 1]$ that controls the pruning aggressiveness. 
Specifically, \citet[Theorem~1]{nikzad2025} consider normalizing constants that take the form of \emph{interval sums} or \emph{intersection sums} of $\pi_i$, defined for node sets $T, U \subseteq N\setminus \{i\}$ as
\bes \label{eq:sums}
	\sum_{T \subseteq J \subseteq U} \pi_i(J)\, \  \ \  \ \text{and} \  \ \sum_{J\subseteq U : J \cap T \neq \emptyset} \pi_i(J)\,,
\ees
respectively. Given $\epsilon \in [0, 1]$, pruning discards some of the possible parent sets so that, when omitting those sets, any interval sum with $|T| \leq 1$ and any intersection sum remain approximations of the exact sum to within a relative error at most $\epsilon$. 

Notably, the node-wise approximations suffice for guaranteeing that the error in the resulting DAG posterior distribution is small: after pruning the scores for all nodes $i \in N$ with $\epsilon := \epsilon'/n$, the total variation distance and the reverse Kullback--Leibler divergence of the pruned DAG posterior from the exact one are bounded by $\epsilon'$ and $\epsilon'/(1-\epsilon')$, respectively \citep[Theorem~2]{nikzad2025}. Since these approximation guarantees concern the DAG posterior itself, they hold regardless of the computational method. In particular, this enables leveraging score pruning also for the blocked Gibbs method, which is not based on computing interval or intersection sums.   

However, methods that do rely on computing interval sums (with $|T| \leq 1$) or intersection sums can also leverage other known techniques to further expedite the computations. One simple and effective technique is to sort the possible parent sets $J$ of node $i$ in decreasing order by the local scores $\pi_i(J)$, as preprocessing, and then for given $T$ and $U$, sum up the scores of parent sets $J$ in this order until the accumulated sum is guaranteed to be within a chosen relative error of the exact sum \citep{Friedman2003, Niinimaki2013}. Also more sophisticated approaches have been studied \citep{Lehtinen2026,Niinimaki2013}. Importantly, these techniques bring a similar speedup for drawing a random parent set from the corresponding approximate distribution. As the moves we introduce in Section~\ref{sec:nm}, as well as the REV and MBR moves employed by Gibby \citep{nikzad2025}, require computing intersection and interval sums and sampling from the corresponding distributions, they directly benefit from these techniques. In contrast, the BG move cannot exploit them.

\section{Generalizing Metropolis--Hastings for Paired Proposals}\label{sec:newacc} 

We next generalize the standard MH framework of mixture proposals that we reviewed in Section~\ref{sec:MCMC}. The key idea is to extend the original mixture into a \emph{double-mixture} by not only selecting one proposal distribution $q_k$, but also pairing it with another distribution~$q_l$. While the candidate for the new state of the chain is still proposed using only $q_k$, the acceptance criterion will also depend on $q_l$. 

\subsection{Duet Sampling}

For a formal treatment, let $\cQ = \{q_k\}_{k \in L}$ be a set of proposal distributions, indexed by labels from a set $L$. For simplicity, assume that $L$ is countable. In the double-mixture, let $\beta_{kl}(G)$ denote the probability of selecting a pair $(q_k, q_l) \in \cQ^2$, allowing dependence on the current state $G$ of the chain. A candidate state $G'$ is proposed with probability $q_k(G'|G)$, and $G'$ is accepted as the new state of the chain with probability
\begin{equation}\label{eq:newacc}
	\alpha^l_k(G'|G)=\min\bigg\{1,\frac{\pi(G')q_l(G|G')\beta_{lk}(G')}{\pi(G)q_k(G'|G)\beta_{kl}(G)}\bigg\} \ .
\end{equation}
As in the standard MH algorithm, if the candidate is rejected, the chain stays at the current state $G$. We call this generalization \emph{duet sampling}, as a \emph{duet} embodies two components working together in coordination. 

With the particular form of the acceptance criterion, it is straightforward to prove that the induced transition kernel satisfies detailed balance.
\begin{theorem}
\label{thm:inv}
The transition kernel obtained by drawing $(q_k,q_l) \in \mathcal{Q}^2$ with probability $\beta_{kl}(G)$, proposing $G'$ with probability $q_{k}(G' | G)$, and accepting $G'$ with probability $\alpha^l_k(G'|G)$ satisfies detailed balance.
\end{theorem}

\begin{proof}
By the definitions, we have 
\bes
	\pi(G)p(G'|G) &=& \sum_{k,l} \pi(G) \beta_{kl}(G) q_k(G'|G) \alpha^{l}_k(G'|G)\,,\\
	\pi(G')p(G|G') &=& \sum_{k,l} \pi(G') \beta_{kl}(G') q_k(G|G') \alpha^{l}_k(G|G')\,.
\ees	

By expanding the acceptance probability terms, the detailed balance condition \eqref{eq:db} can be written as 
\begin{equation}\label{eq:d2}
\begin{split}
	 &\,\, \sum_{k,l} \min\Big\{\pi(G)\,\beta_{kl}(G) q_k(G'|G)\,,\, \pi(G')\beta_{lk}(G')\,q_l(G|G')\Big\} \\ 
	=&\,\, \sum_{{k,l}}\min\Big\{\pi(G')\beta_{kl}(G')q_k(G|G')\,,\, \pi(G)\beta_{lk}(G)q_{l}(G'|G)\Big\}\,. 
\end{split}
\end{equation}
This is seen to hold by swapping (renaming) the labels $k$ and $l$ in the latter summation. 
\end{proof}

Worth noting, duet sampling can also be viewed as an instantiation of \emph{involutive MCMC} \citep[iMCMC,][]{Neklyudov2020} and the broader \emph{generalized auxiliary-variable Metropolis–Hastings} framework of \cite{storvik2011}, both of which unify various known algorithms for constructing reversible Markov chains on discrete, continuous, and mixed state spaces. We discuss the relationship of duet sampling and these frameworks in detail in Appendix~\ref{sec:imcmc}.

When applying duet sampling in a concrete setting, one has to specify the mixture probabilities $\beta_{kl}(G)$. We will refer to a probability distribution $\{\beta_{kl}\}$ on $L^2$, understood as functions of the state~$G$, as a \emph{pairing} on the set of proposal distributions $\{q_k\}$. We will see in Section~\ref{sec:nm} that the dependence on $G$ is necessary for the EC and AC moves, but not for the USR move.

\subsection{How to Construct Duet Samplers}

An inappropriate choice of pairings can cause the acceptance probability \eqref{eq:newacc} to vanish, resulting in automatic rejection of the proposal. Thus, designing pairings that guarantee a nonzero acceptance probability is desirable. Since the denominator in \eqref{eq:newacc} is necessarily nonzero,  as $G'$ is proposed using $\beta_{kl}(G)$ and $q_k(G' | G)$, our focus is therefore on ensuring that the numerator is nonzero. We formalize this requirement as follows: 
\begin{definition}[Effective pairing]
A pairing $\{\beta_{kl}\}$ on $\{q_k\}$ is \emph{effective} if, for all $k$, $l$, $G$, and $G'$,  
\bes 
	\beta_{kl}(G) q_k(G'| G)>0\, \text{ implies } \,\beta_{lk}(G') q_{l}(G| G') > 0\, .
\ees
\end{definition}

How can one construct effective pairings? To address this question, we consider sets of proposal distributions $\mathcal{Q}$ that satisfy at least one of two modest conditions. We first introduce these two conditions and then identify the respective pairing templates that guarantee effectiveness. 

The first condition on $\mathcal{Q}$ is stronger, requiring that every transition induced by $q_k \in \mathcal{Q}$ can be reversed by \emph{any} $q_l \in \mathcal{Q}$.
\begin{definition}[Strong positivity]
A set of proposal distributions $\{q_k\}$ is \emph{strongly positive} if for all $k$, $l$, $G$, and $G'$, 
\bes 
	q_k(G'| G)>0\, \text{ implies }\,  q_{l}(G| G') > 0 \,.
\ees
\end{definition}

The second condition is weaker, only requiring that every transition induced by $q_k \in \mathcal{Q}$ can be reversed by \emph{some} $q_l \in \mathcal{Q}$. However, as becomes evident later, it is useful to require that we can map $k$ to $l$ by an involution, i.e., a bijection $f : L \rightarrow L$ satisfying $f(f(k)) = k$.
\begin{definition}[$f$-positivity]
A set of proposal distributions $\{q_k\}$ is \emph{positive under} an involution $f$ (or $f$\emph{-positive}) if for all $k$, $G$, and $G'$,
\bes 
	q_k(G'| G)>0\, \text{ implies }\,  q_{f(k)}(G| G') > 0 \, .
\ees
\end{definition}
Note that strong positivity implies $f$-positivity under any involution $f$.

We now describe two simple design templates that yield effective pairings when $\mathcal{Q}$ is strongly positive or $f$-positive. The first template is tailored for strongly positive $\mathcal{Q}$. It simply draws that the proposal distributions $q_k$ and $q_l$ independently of each other. Let $\beta_k(G)$ denote the marginal probability $\sum_l \beta_{kl}(G)$ for all $k$ and~$G$. 
\begin{definition}[Independent pairing]
A pairing $\{\beta_{kl}\}$ on $\{q_k\}$ is \emph{independent} if
\bes 
	\beta_{kl}(G)=\beta_k(G)\beta_l(G) 
	\quad
	\text{for all $k$, $l$, and $G$}.
\ees
\end{definition}

\begin{proposition}\label{prop:eff1}
Any strongly positive $\{q_k\}$ admits effective independent pairings $\{\beta_{kl}\}$.
\end{proposition}
\begin{proof}
Choose any independent pairing such that for any $k$, $l$, $G$, and~$G'$,
\bes 
	 \beta_k(G)\beta_l(G)q_k(G'| G)>0 \, \text{ implies } \, \beta_{k}(G')\beta_{l}(G')>0  \,.
\ees
Such pairings clearly exist, for we may put, e.g., $\beta_k(G) = \beta_k(G')$ for all $k$, $G$, and $G'$. 

Suppose $\{q_k\}$ is strongly positive. Then $q_k(G'| G)>0$ implies $q_l(G| G')>0$. Taken together, these properties imply the condition of effectiveness. 
%Applying this pairing to a strongly positive $\mathcal{Q}$ ensures that the numerator of the acceptance ratio~\eqref{eq:newacc} is always positive, %establishing effectiveness.
\end{proof}

Our second template is designed for sets $\mathcal{Q}$ that are positive under some involution $f$. Now we draw $q_k \in \cQ$ with probability $\beta_k(G)$ and deterministically assign $q_l$ by $l := f(k)$. 
\begin{definition}[Involutive pairing]
A pairing $\{\beta_{kl}\}$ on $\{q_k\}$ is said to be \emph{involutive under}~$f$ (or an $f$\emph{-pairing}) if $f$ is an involution and
\bes 
	\beta_{k f(k)}(G)=\beta_k(G)
	\quad
	\text{for all $k$ and $G$}\,. 
\ees
\end{definition}
Note that in an involutive pairing, the probability $\beta_{kl}(G)$ must vanish if $l \neq f(k)$, since $f$ is a bijection and $\sum_k \beta_k(G) = 1$. Furthermore, if $f$ were not an involution, then $f(k) = l$ but $f(l) \neq k$ for some $k$ and $l$. Consequently, $\beta_{lk}(G') = 0$ in \eqref{eq:newacc}, rendering the acceptance probability zero. Thus, the requirement that $f$ is an involution is a necessary condition for a deterministic pairing to be effective.

\begin{proposition}\label{prop:eff2}
Any $f$-positive $\{q_k\}$ admits effective $f$-pairings $\{\beta_{kl}\}$.
\end{proposition}
\begin{proof}
Choose any $f$-pairing such that for any $k$, $G$, and~$G'$,
\bes 
	 \beta_k(G)q_k(G'| G)>0 \, \text{ implies } \, \beta_{f(k)}(G')>0  \,.
\ees
Such pairings clearly exist, for we may put, e.g., $\beta_k(G) = \beta_{f(k)}(G')$ for all $k$, $G$, and $G'$. 

Suppose $\{q_k\}$ is $f$-positive. Then $q_k(G'| G)>0$ implies $q_{f(k)}(G| G')>0$. Taken together, these properties imply the condition of effectiveness. 
%When $\mathcal{Q}$ is positive under the same involution $f$, the numerator of the acceptance ratio~\eqref{eq:newacc} is always positive, establishing effectiveness.
\end{proof}

In some cases, such as in the edge-constrained and ancestor-constrained moves that we will introduce in the next section, the set $\mathcal{Q}$ is $f$-positive with respect to a unique involution $f$. In contrast, for strongly positive $\cQ$ we can design effective involutive pairings for each involution $f: L \rightarrow L$, in addition to effective independent pairings. This provides us with complete freedom on the choice of the involution, allowing us to select one that enhances the mixing properties of the Markov chain. As we shall see, this flexibility is exploited by both the Markov blanket resampling move and the newly proposed unconstrained sequential resampling move detailed in the next section.

Finally, we observe a seemingly counterintuitive property. For both independent and involutive pairings, the transition kernel $p_{kl}(G' | G) = q_k(G' | G)\,\alpha_k^l(G' | G)$ induced by a single move (i.e., component of the double-mixture) does not, in general, satisfy detailed balance. Indeed, we see from \eqref{eq:d2} that detailed balance holds only after summing over $k$ and $l$, not per summand. For a single move, detailed balance only holds when $l = k$, which corresponds to the trivial involutive pairing where the involution is the identity map. This specific case falls within the standard Metropolis--Hastings framework.

\section{New Moves}\label{sec:nm}

Leveraging the duet sampling framework introduced in Section~\ref{sec:newacc}, we next introduce three new moves that transition from one DAG to another by resampling the parent sets of two or more nodes.

\subsection{The Edge-Constrained Move}\label{sec:ecpr}

Suppose the current DAG is $G$. The \emph{edge-constrained} (EC) move proposes a new DAG $G'$ by first drawing a node pair $ij$ with probability $\gamma_{ij}$, and then sampling new parent sets $P'_i$ and $P'_j$ for $i$ and $j$, respectively. These sets are subject to constraints that depend on whether $ij$ or $ji$ is an edge in $G$, as follows:
\begin{description} 
\item[]\emph{Add} $ij$: if neither $ji$ nor $ij$ is an edge in $G$, then require $P'_j \ni i$. 
\item[]\emph{Remove} $ij$: if $ij$ is an edge in $G$, then require $P'_j \not\ni i$ and $P'_i \not\ni j$.
\item[]\emph{Reverse to} $ij$: if $ji$ is an edge in $G$, then require $P'_j \ni i$.
\end{description}
Thus, these constraints, we denote by $c$, are fully determined by $G$ and the nodes $i$ and $j$; we keep these dependencies implicit in the notation. When it is necessary to explicit the constraint type, we write $c = \text{add}$, $c = \text{rem}$, or $c = \text{rev}$. 

\subsubsection{Proposal Distribution}

Once the constraint $c$ has been identified, the nodes $i$ and $j$ are orphaned, yielding an intermediate DAG $G_0$. Then, the candidate parent sets $P'_i$ and $P'_j$ are sampled sequentially from those that preserve acyclicity and satisfy the constraint $c$. 

When $c=\text{add}$ or $c=\text{rev}$, the sampling order does not affect the joint proposal distribution, as the ancestor relationship between $i$ and $j$, which determines their candidate parents, are already determined by $c$. Indeed, if $c$ forces $i$ to be a parent of $j$ then the descendants of $j$ in the intermediate DAG cannot be candidate parents for $i$. 

In contrast, when $c=\text{rem}$, the ancestor relationship between $i$ and $j$ in $G'$ is not known a priori. Thus, the sampling orders $(i,j)$ and $(j,i)$ can possibly generate different candidate parent sets for these nodes. We prefer the order $(i,j)$ because it results in a higher chance of reversing the ancestor relationships between $i$ and the descendants of $j$ in $G$, which is desirable for improving exploration of the DAG posterior. As discussed in Section \ref{sec:pmt}, favoring such reversals is beneficial because, when resampling the parent set of a node $i$, its descendants are excluded from the candidate set to prevent cycles. Thus, the nature of the move makes it difficult to reverse existing ancestor relationships, and design choices that facilitate such reversals become particularly valuable.

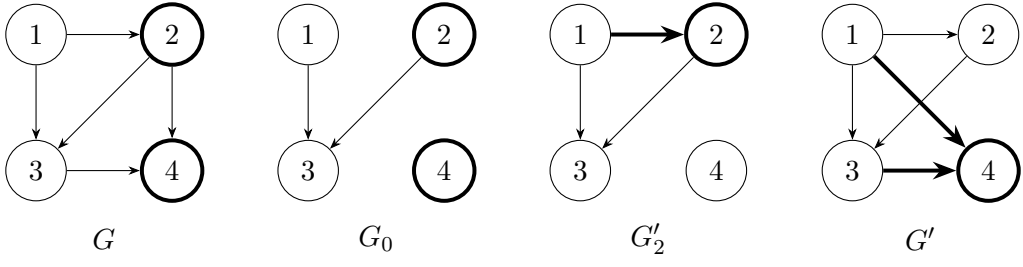
\begin{figure}[t]
\centering
\begin{tikzpicture}[->, >={Stealth[scale=0.9]}, node distance=2cm, scale=0.9]
\tikzstyle{node}=[circle, draw, minimum size=8mm]

% --- DAG 1 ---
\begin{scope}
\node[node] (A1) at (0,0) {1};
\node[node, line width=1.5pt] (B1) at (2,0) {2};
\node[node, line width=1.5pt] (D1) at (2,-2) {4};
\node[node] (C1) at (0,-2) {3};

\draw (A1) -> (B1);
\draw (B1) -> (D1);
\draw (B1) -> (C1);
\draw (A1) -> (C1);
\draw (C1) -> (D1);

\node at (1,-3) {$G$};
\end{scope}

% --- DAG 2 ---
\begin{scope}[xshift=4cm]
\node[node] (A2) at (0,0) {1};
\node[node, line width=1.5pt] (B2) at (2,0) {2};
\node[node, line width=1.5pt] (D2) at (2,-2) {4};
\node[node] (C2) at (0,-2) {3};

\draw (A2) -> (C2);
\draw (B2) -> (C2);

\node at (1,-3) {$G_0$};
\end{scope}

% --- DAG 3 ---
\begin{scope}[xshift=8cm]
\node[node] (A3) at (0,0) {1};
\node[node, line width=1.5pt] (B3) at (2,0) {2};
\node[node] (D3) at (2,-2) {4};
\node[node] (C3) at (0,-2) {3};

\draw[line width=1.5pt] (A3) -> (B3);
\draw (A3) -> (C3);
\draw (B3) -> (C3);

\node at (1,-3) {$G'_2$};
\end{scope}

% --- DAG 4 ---
\begin{scope}[xshift=12cm]
\node[node] (A4) at (0,0) {1};
\node[node] (B4) at (2,0) {2};
\node[node, line width=1.5pt] (D4) at (2,-2) {4};
\node[node] (C4) at (0,-2) {3};

\draw (A4) -> (C4);
\draw[line width=1.5pt] (A4) -> (D4);
\draw[line width=1.5pt] (C4) -> (D4);
\draw (A4) -> (B4);
\draw (B4) -> (C4);

\node at (1,-3) {$G'$};
\end{scope}

\end{tikzpicture}

\caption{Illustration of an EC move transforming the current DAG $G$ into the proposal~$G'$. First, a node pair is sampled. In this example the pair is $(2,4)$. Since there is a directed edge from $2$ to $4$, the corresponding edge constraint is $c = \mathrm{rem}$. The nodes $2$ and $4$ are then orphaned, yielding the intermediate DAG $G_0$. Next, a new parent set $P'_2$ is sampled from those that ensure acyclicity and satisfy the edge constraint $c$ (in this case, node 4 must not be a parent of 2), producing $G'_2$. Finally, a new parent set $ P'_{\text{4}} $ is sampled from those that ensure acyclicity  and satisfy the edge constraint $ c $ (node 2 must not be a parent of 4), generating the proposal $G'$.}
\label{fig:ec}
\end{figure}

Following the order $(i,j)$, the proposal distribution associated with the node pair $ij$ and the constraint $c$ becomes
\be\label{eq:q1}
    q^{ij}_c(G'|G) = \frac{ \pi_i(P'_i)\, \pi_j(P'_j)}{Z_i^{c}(G_0) \,Z_j^{c}(G'_i)}\,,
\ee
where the DAG $G'_i$ is obtained by incorporating the new parent set $P'_i$ into $G_0$. The normalizing constant $Z_i^{c}(G_0)$, which we introduced in its general form in \eqref{eq:q0}, is the sum of the scores of all parent sets of $i$ that do not create  cycles when added to the DAG $G_0$, and that satisfy the edge constraint $c$. The term $Z_j^{c}(G'_i)$ is defined analogously for node $j$ with respect to $G'_i$. Figure \ref{fig:ec} illustrates how the EC move proposes a transition from $G$ to $G'$.

Importantly, the normalizing constants $Z_i^{c}(G_0)$ and $Z_j^{c}(G'_i)$ can be written as interval sums of $\pi_i$ and $\pi_j$ respectively; see Appendix~\ref{sec:norm} for details. Thus, we can apply the techniques mentioned in Section \ref{sec:pru} to speed up the computation of the terms and sampling from the corresponding distributions.

\subsubsection{Involutive Pairing and Acceptance Ratio}

We leverage the generalized acceptance probability $\alpha_k^l$ defined in \eqref{eq:newacc} to guarantee detailed balance. Specifically, we next give an involution $f$ such that the set $\cQ$ of proposal distributions of the EC move is $f$\emph{-positive}. This allows us to construct an \emph{effective involutive pairing} and thereby guarantee a nonzero acceptance probability. 

As there are three possible constraints for each node pair $ij$, the total  number of proposal distributions is $|\mathcal{Q}|=3n(n-1)$. For notational convenience, we define the involution $f$ on the set $\cQ$ rather on the set of labels. For each $ij$, let 
\bes 
	f(q^{ij}_{\text{add}})=q^{ij}_{\text{rem}}\,,
	\quad f(q^{ij}_{\text{rem}})=q^{ij}_{\text{add}}\,,\quad \text{and} 
	\quad f(q^{ij}_{\text{rev}})=q^{ji}_{\text{rev}}\,. 
\ees
\begin{lemma}\label{lemma:ec}
The set $\{q_c^{ij}\}$ is positive under $f$.
\end{lemma}
\begin{proof}
The forward transition induced by $q_{ij}^c$ from $G$ to $G'$ and its resulting backward transition via $f(q_c^{ij})$ from $G'$ share the same orphaned graph $G_0$ for all $G$, $G'$, $ij$, and $c$. Furthermore, for all $ij$ and $c$, the edge constraint of $q_c^{ij}$ is equivalent to the condition of the edge constraint of $f(q_c^{ij})$ and vice versa. Therefore, in the backward transition from $G'$ via $f(q_c^{ij})$, we can incorporate $P_i$ and $P_j$ in $G_0$ with positive probability as no cycle is introduced in the process and the corresponding edge constraint is satisfied. Hence, the set $\{q_c^{ij}\}$ is positive under $f$. 
\end{proof}

Now, by Proposition~\ref{prop:eff2}, the set $\mathcal{Q}$ admits effective $f$-pairings. However, we have already fixed the pairing up to the values of the scalars $\gamma_{ij}$: the probability $\beta_c^{ij}(G)$ of sampling $q_c^{ij} \in \mathcal{Q}$ equals the probability $\gamma_{ij}$ of sampling the node pair $ij$ for the unique constraint~$c$ whose conditions are satisfied in $G$, and $\beta_c^{ij}(G)=0$ for the other two constraints~$c$. To see that this particular pairing is effective, observe that  
\bes
\begin{aligned}
q_{\mathrm{add}}^{ij}(G'|G)>0
\quad&\text{implies}\quad \beta_{\mathrm{rem}}^{ij}(G')=\gamma_{ij}\,\\
q_{\mathrm{rem}}^{ij}(G'|G)>0
\quad&\text{implies}\quad \beta_{\mathrm{add}}^{ij}(G')=\gamma_{ij}\,, \; \text{and}\\
q_{\mathrm{rev}}^{ij}(G'|G)>0
\quad&\text{implies}\quad \beta_{\mathrm{rev}}^{ji}(G')=\gamma_{ji}\,.
\end{aligned}
\ees
Thus, the acceptance probability \eqref{eq:newacc} reduces to $\min\{1, r\}$, where for $c = \text{add}$, $c = \textrm{rem}$, and $c = \text{rev}$, we have respectively 
\bes
	r = \frac{Z_i^{\text{add}}(G_0)\, Z_j^{\text{add}}(G'_i)}{Z_i^{\text{rem}}(G_0)\, Z_j^{\text{rem}}(G_i)}\,,\quad
	r = \frac{Z_i^{\text{rem}}(G_0)\, Z_j^{\text{rem}}(G'_i)}{Z_i^{\text{add}}(G_0)\, Z_j^{\text{add}}(G_i)}\,,\quad \textrm{and}\quad
	r = \frac{\gamma_{ji}\, Z_i^{\text{rev}}(G_0)\, Z_j^{\text{rev}}(G'_i)}{\gamma_{ij}\, Z_j^{\text{rev}}(G_0)\, Z_i^{\text{rev}}(G_j)}\,.
\ees
We have shown the following:
\begin{proposition}\label{prop:ec2}
The $f$-pairing above is effective, provided that $\gamma_{ij} = 0$ whenever $\gamma_{ji} = 0$.
\end{proposition}
In the experiments (Section \ref{sec:exp}), we set $\gamma_{ij}=\gamma=(n(n-1))^{-1}$ for all node pairs $ij$ ($i \neq j$).

\subsubsection{Comparison with Related Previous Moves}\label{sec:eccomp}

The EC move can be viewed as a composition of the edge-reversal (REV) move of \cite{Grzegorczyk2008} and the single-edge moves (GC) of  \cite{Giudici2003}, which propose a DAG by adding, removing or reversing a single edge. More specifically, REV is a special case of EC, as its proposal distribution is equal to that of EC under the reverse-to $ij$ constraint, multiplied by $|E|/|E'|$, where $E$ and $E'$ denote the edge sets of $G$ and $G'$, respectively. By augmenting the REV move with the GC's single-edge dynamics, we allow fast transitions among the three possible configurations of a node pair $ij$: an edge in either direction or no edge, without systematically favoring any particular configuration.

Unlike GC, EC retains REV's ability to resample the parent sets of $i$ and $j$, enabling transitions to a substantially broader set of DAGs. This facilitates more effective exploration of the DAG posterior, whereas GC can propose at most $n(n-1)$ DAGs from a given state, thereby precluding longer jumps to high-probability regions.

\subsection{The Ancestor-Constrained Move}\label{sec:acpr}

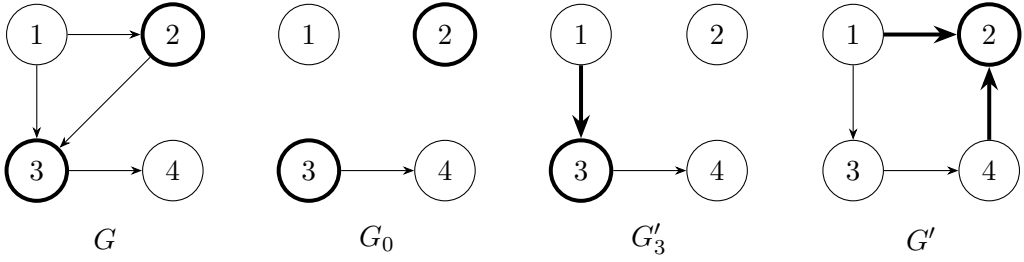
\begin{figure}[t]
\centering
\begin{tikzpicture}[->, >={Stealth[scale=0.9]}, node distance=2cm, scale=0.9]
\tikzstyle{node}=[circle, draw, minimum size=8mm]

% --- DAG 1 ---
\begin{scope}
\node[node] (A1) at (0,0) {1};
\node[node, line width=1.5pt] (B1) at (2,0) {2};
\node[node] (D1) at (2,-2) {4};
\node[node, line width=1.5pt] (C1) at (0,-2) {3};

\draw (A1) -> (B1);
\draw (B1) -> (C1);
\draw (A1) -> (C1);
\draw (C1) -> (D1);

\node at (1,-3) {$G$};
\end{scope}

% --- DAG 2 ---
\begin{scope}[xshift=4cm]
\node[node] (A2) at (0,0) {1};
\node[node, line width=1.5pt] (B2) at (2,0) {2};
\node[node] (D2) at (2,-2) {4};
\node[node, line width=1.5pt] (C2) at (0,-2) {3};

\draw (C2) -> (D2);

\node at (1,-3) {$G_0$};
\end{scope}

% --- DAG 3 ---
\begin{scope}[xshift=8cm]
\node[node] (A3) at (0,0) {1};
\node[node] (B3) at (2,0) {2};
\node[node] (D3) at (2,-2) {4};
\node[node, line width=1.5pt] (C3) at (0,-2) {3};

\draw (C3) -> (D3);
\draw[line width=1.5pt] (A3) -> (C3);

\node at (1,-3) {$G'_3$};
\end{scope}

% --- DAG 4 ---
\begin{scope}[xshift=12cm]
\node[node] (A4) at (0,0) {1};
\node[node, line width=1.5pt] (B4) at (2,0) {2};
\node[node] (D4) at (2,-2) {4};
\node[node] (C4) at (0,-2) {3};

\draw (A4) -> (C4);
\draw[line width=1.5pt] (A4) -> (B4);
\draw[line width=1.5pt] (D4) -> (B4);
\draw (C4) -> (D4);

\node at (1,-3) {$G'$};
\end{scope}

\end{tikzpicture}

\caption{Illustration of an AC move transforming the current DAG $G$ into the proposal~$G'$. First, a node pair is sampled. In this example, the pair is $( \text{3}, \text{2} )$. Since $ \text{2} $ is an ancestor of $ \text{3} $ in $G_0$, the corresponding ancestor constraint is $ c = \text{rev} $, thus the move forces $ \text{3} $ to be an ancestor of $ \text{2} $ in $G'$. The nodes $ \text{3} $ and $ \text{2} $ are then orphaned, yielding the intermediate DAG $ G_0 $. Next, a new parent set $ P'_3 $ is sampled from those that maintain acyclicity and satisfy the ancestor constraint $c$ (in this case, node 2 and its descendants in $G_0$ must not be parents of $3$). This produces the DAG $G'_3$. Finally, a new parent set $ P'_2 $ is sampled from those that guarantee acyclicity and satisfy the ancestor constraint $c$ (node 3 must be an ancestor of 2), generating the DAG~$G'$. }
\label{fig:AC}
\end{figure}

The \emph{ancestor-constrained} (AC) move follows the same ideas of the EC move, with the only difference that the constraints concern the ancestor relationship between the selected nodes $i$ and $j$, not the adjacency relationship. Let us write $i \rightsquigarrow j$ when $i$ is an ancestor of $j$ in the graph in question. The constraints between $i$ and $j$ are expressed as follows:
\begin{itemize}
    \item[] \emph{Add $i\rightsquigarrow j$}. If  $i \not\rightsquigarrow j$ and $j \not\rightsquigarrow i$ in $G$, then require  $i \rightsquigarrow j$ in $G'$. 
    \item[] \emph{Remove $i\rightsquigarrow j$}. If $i \rightsquigarrow j$ in $G$, then require $i \not\rightsquigarrow j$ and $j \not\rightsquigarrow i$ in~$G'$.
    \item[] \emph{Reverse to $i\rightsquigarrow j$}. If $j \rightsquigarrow i$ in $G$, then require $i \rightsquigarrow j$ in $G'$.
\end{itemize}
Thus, the constraints are determined by the ancestor relationship between nodes $i$ and $j$ in the DAG $G$. 

The sampling order of the new parent sets does not affect the possible parent sets for nodes $i$ and~$j$, since the ancestor relations in $G'$ are already determined by the ancestor constraints shown above. The proposal distribution is the same as \eqref{eq:q1} although the corresponding constraints differ. Figure~\ref{fig:AC} illustrates how the AC move generates a proposal~$G'$. 

Lemma~\ref{lemma:ec}, which establishes the $f$-positivity of the set of proposal distribution of the EC move, also applies to the AC move. Consequently, Proposition~\ref{prop:ec2}, which guarantees effectiveness, extends directly to AC.

It is not difficult to see that the normalizing constants of the AC move can be written as intersection and interval sums; see Appendix~\ref{sec:norm} for details. Thus, we may leverage the techniques mentioned in Section~\ref{sec:pru} to speed up the computations. 

\subsection{The Unconstrained Sequential Resampling Move}\label{sec:bmh}

The \emph{unconstrained sequential resampling} (USR) move is designed to overcome the high computational cost of the blocked Gibbs (BG) move of \citet{Goudie2016}, while retaining the ability to resample a block of parent sets without imposing constraints other than acyclicity. By sampling the new parent sets sequentially rather than jointly, USR can generate a DAG proposal much faster than BG. However, the gained efficiency comes at the expense of losing the 100\,\% acceptance rate of Gibbs sampling. We turn this seeming shortcoming into an advantage by leveraging the flexibility of duet sampling to construct chains with favorable mixing properties. Indeed, it is known that suitably designed Metropolis--Hastings type algorithms can, in general, improve upon Gibbs sampling by reducing autocorrelation and asymptotic variance \citep{neal1998, peskun1973}. 

\subsubsection{Proposal Distribution}

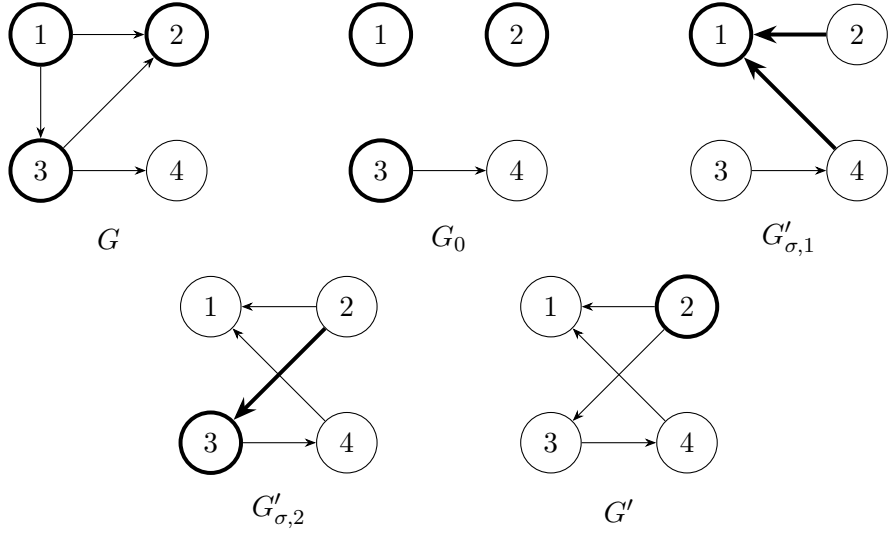
\begin{figure}[t!]
\centering
\begin{tikzpicture}[->, >={Stealth[scale=0.9]}, node distance=2cm, scale=0.9]
\tikzstyle{node}=[circle, draw, minimum size=8mm]

% ---------------- First row ----------------
% DAG 1
\begin{scope}
\node[node, line width=1.5pt] (A1) at (0,0) {1};
\node[node, line width=1.5pt] (B1) at (2,0) {2};
\node[node, line width=1.5pt] (C1) at (0,-2) {3};
\node[node] (D1) at (2,-2) {4};

\draw (A1) -> (B1);
\draw (A1) -> (C1);
\draw (C1) -> (B1);
\draw (C1) -> (D1);

\node at (1,-3) {$G$};
\end{scope}

% DAG 2
\begin{scope}[xshift=5cm]
\node[node, line width=1.5pt] (A2) at (0,0) {1};
\node[node, line width=1.5pt] (B2) at (2,0) {2};
\node[node, line width=1.5pt] (C2) at (0,-2) {3};
\node[node] (D2) at (2,-2) {4};

\draw (C2) -> (D2);

\node at (1,-3) {$G_0$};
\end{scope}

% DAG 3
\begin{scope}[xshift=10cm]
\node[node, line width=1.5pt] (A3) at (0,0) {1};
\node[node] (B3) at (2,0) {2};
\node[node] (C3) at (0,-2) {3};
\node[node] (D3) at (2,-2) {4};

\draw[line width=1.5pt] (B3) -> (A3);
\draw[line width=1.5pt] (D3) -> (A3);
\draw (C3) -> (D3);

\node at (1,-3) {$G'_{\sigma,1}$};
\end{scope}

% ---------------- Second row ----------------
% DAG 4
\begin{scope}[xshift=2.5cm, yshift=-4cm]
\node[node] (A4) at (0,0) {1};
\node[node] (B4) at (2,0) {2};
\node[node, line width=1.5pt] (C4) at (0,-2) {3};
\node[node] (D4) at (2,-2) {4};

\draw (B4) -> (A4);
\draw (D4) -> (A4);
\draw[line width=1.5pt] (B4) -> (C4);
\draw (C4) -> (D4);

\node at (1,-3) {$G'_{\sigma,2}$};
\end{scope}

% DAG 5
\begin{scope}[xshift=7.5cm, yshift=-4cm]
\node[node] (A5) at (0,0) {1};
\node[node, line width=1.5pt] (B5) at (2,0) {2};
\node[node] (C5) at (0,-2) {3};
\node[node] (D5) at (2,-2) {4};

\draw (B5) -> (A5);
\draw (B5) -> (C5);
\draw (D5) -> (A5);
\draw (C5) -> (D5);

\node at (1,-3) {$G'$};
\end{scope}

\end{tikzpicture}
\caption{Illustration of a USR move transforming the current DAG $G$ into the proposal~$G'$. First, a permutation $\sigma$ is sampled. Here $\sigma=(1,3,2)$. The nodes in $\sigma$ are orphaned, yielding the intermediate DAG $ G_0 $. Next, the new parent sets of the nodes in $\sigma$ are sequentially sampled from those that preserve acyclicity, according to the order given by $\sigma$, and incorporated into $G_0$. In detail, we obtain the DAG~$G'_{\sigma,1}$ after sampling $P'_1$, the DAG $G'_{\sigma,2}$ after sampling $P'_3$, and finally $G'$ after sampling $P'_2$. In this particular case, the only candidate parent set of node~2 is the empty parent set as it is the only one that preserves acyclicity.}
\label{fig:bmh}
\end{figure}

The USR move first selects a random node subset $S$, with the block size $|S|$ specified as an input parameter. Then a permutation~$\sigma$ of the nodes in $S$ is sampled with probability $\beta_\sigma$, independently of the current state $G$. Next, following the procedure detailed in Section~\ref{sec:pmt}, the DAG $G_0$ is obtained by orphaning the nodes in $S$, and a sequence of intermediate DAGs are constructed as follows. Starting from $G_0$, for each node in $S$ in the order induced by $\sigma$, a new parent set is sampled with probability proportional to its score and incorporated into the intermediate DAG at hand. No edge or ancestor constraints are enforced; the only restriction is that the proposed parent set must not introduce cycles in the intermediate DAGs.
Following the notation used in Section~\ref{sec:pmt}, the proposal distribution takes the form
\be\label{eq:q3}
	q_\sigma(G'| G)
	= \frac{\prod_{s=1}^{|S|}\pi_{\sigma(s)}(P'_{\sigma(s)})}{Z^\sigma(G')}\,, 
	\quad \text{where} \quad 
	Z^\sigma(G')
	= \prod_{s=1}^{|S|} Z_{\sigma(s)}(G'_{\sigma,s-1})\,.
\ee
Recall that $G'_{\sigma, s-1}$ is the DAG obtained by adding the parent sets $P'_{\sigma(1)}, P'_{\sigma(2)},\ldots, P'_{\sigma(s-1)}$ to the orphaned DAG $G_0$. The term $Z_{\sigma(s)}(G'_{\sigma, s-1})$ is the sum of the scores of all possible parents sets of node $\sigma(s)$ that would not create a cycle when incorporated to the DAG~$G'_{\sigma, s-1}$. In the notation, we do not include any additional move label ``$m$'', since the proposal distribution is completely determined by $\sigma$. Figure~\ref{fig:bmh} illustrates how the USR move proposes a new DAG $G'$ from $G$.

Since no constraints beyond acyclicity is imposed on the new parent sets, each normalizing term $Z_{\sigma(s)}(G'_{\sigma, s-1})$ is an interval sum with $T=\emptyset$. Thus, the techniques mentioned in Section~\ref{sec:pru} can be leveraged for computational speedups.

The computational cost of the total normalizing constant $Z^\sigma(G')$ scales linearly with the size of the set $S$, making the computations feasible even for large $S$. However, large sets~$S$ are not necessarily desirable. Namely, there may be high-probability DAGs $G'$ that can be proposed with non-negligible probability only under node orderings that match~\footnote{\label{fn:ordering}We say that an ordering $\sigma$ on $S$ \emph{matches} an ordering $\rho$ on $V \supseteq S$ if the ordering of the nodes in $\sigma$ equals their relative ordering in $\rho$, that is, $\sigma$ is a restriction of $\rho$ to $S$. When selecting the parent set of a node in $\sigma$, the subsequent nodes are available as candidate parents, whereas the predecessors might not be. By sequentially integrating the parent sets of the nodes in $\sigma$ into the intermediate DAG, we progressively reduce the pool of candidate parents for subsequent nodes. This reduction is maximized when $\sigma$ matches a reverse topological ordering of $G'$, making such an ordering the most favorable for proposing $G'$.} 
some reverse topological ordering of $G'$. When $|S|$ is small, the probability of sampling a matching node ordering is relatively high, whereas for large $|S|$ the chances are small, the number of all possible node orderings scaling as $|S|!$. Moreover, the larger the size $|S|$, the larger the risk of disrupting high-scoring substructures of the current DAG, that may be difficult to recover using a non-matching node ordering. On the other hand, small~$|S|$ restricts the set of DAGs that can be proposed from the current DAG $G$, making the Markov chain more likely to get trapped in a local mode. Consequently, choosing the block size involves a trade-off between enabling long jumps in the DAG space and avoiding the proposal of low-probability DAGs.

\subsubsection{Move Variants Through Independent and Involutive Pairings}\label{sec:accbmh}

To complete the specification of the USR move, we select an appropriate pairing of the proposal distributions $q_\sigma$. It will be convenient to consider pairings in the set of permutations on $S$, which we denote by $\perms{S}$. While pairing a permutation deterministically with itself would amount to using the standard MH algorithm, duet sampling enables also other constructions, as described in Section~\ref{sec:newacc}. 

Before selecting the suitable pairings, we establish strong positivity of the set~$\{q_\sigma\}$.
\begin{lemma}\label{lemma:usr}
The set $\{q_\sigma\}_{\sigma\in\perms{S}}$ is strongly positive.
\end{lemma}
\begin{proof}
When $q_\sigma$ with $\sigma \in \perms{S}$ is used to propose $G'$ from $G$, the orphaned graph $G_0$ is identical to that obtained by performing a USR move from $G'$ using any permutation $\tau\in\perms{S}$. Thus, in the backward transition from $G'$ via any $q_\tau$ with $\tau \in \perms{S}$, we can iteratively incorporate the parent sets $\{P_i\}_{i\in S}$ in the intermediate DAGs with positive probability, as no cycle is generated in the process.  This guarantees strong positivity of $\{q_\sigma\}_{\sigma\in\perms{S}}$.
\end{proof}

This fact allows us to construct variants of the USR move that exploit either an independent or involutive pairing. Each variant is specified by the probabilities $\beta_{\sigma\tau}$ of selecting the pair $(\sigma, \tau) \in \perms{S}^2$; for simplicity, we let these probabilities be independent of the current state~$G$.  The acceptance probability of the generic USR move (per mixture component) becomes 
\begin{equation}\label{eq:a2}
    \alpha^\tau_\sigma(G'|G)=\min\bigg\{1\,,\frac{\beta_{\tau\sigma}\, Z^\sigma(G')}{\beta_{\sigma\tau}\, Z^\tau(G)}\bigg\}\,.
\end{equation}

For an independent pairing, we put $\beta_{\sigma\tau}=\beta_{\tau\sigma}=\beta_\sigma\beta_\tau$ with some probability distribution $(\beta_\sigma)$ on $\perms{S}$. We will refer to this as a USR-\emph{rnd} move. Note that the terms $\beta_{\sigma \tau}$ and $\beta_{\tau \sigma}$ cancel each other in the ratio \eqref{eq:a2}.

For an involutive pairing, we can freely select the involution $f: \perms{S}\to\perms{S}$ to couple~$\sigma$ and~$\tau$. We consider two specific involutions: the identity map $f(\sigma) = \sigma$ and the reversal map $f(\sigma) = \sigma^\text{rev}$, where $\sigma^\text{rev}$ is obtained by reversing the order of elements in $\sigma$. We refer to the resulting two moves as USR-\emph{id} and USR-\emph{rev}, respectively. Note that USR-id is the variant one would obtain using the standard MH algorithm.

Recall that strong positivity implies $f$-positivity under any involution $f$. By Propositions~\ref{prop:eff1} and \ref{prop:eff2}, we get the following: 
\begin{proposition}\label{prop:usr2}
The pairings in the USR-rnd move are effective.
\end{proposition}
\begin{proposition}\label{prop:usr3}
The pairings in the USR-rev move are effective, provided that $\beta_{\sigma^\mathrm{rev}} = 0$ whenever $\beta_{\sigma} = 0$.
\end{proposition}
For simplicity, in the experiments reported in Section~\ref{sec:exp}, we set $\beta_\sigma=\beta=1/|S|!$ for all $\sigma$, consistently across all the three USR move variants.

The USR-id is inferior to USR-rev and USR-rnd. The acceptance rate of USR-id is low for DAG proposals that reverse the ancestor relationship between nodes, causing the Markov chain to get stuck in regions of the posterior containing DAGs with similar ancestor relationships. Conversely, USR-rev promotes longer jumps in the DAG space by favoring proposals that reverse ancestor relationships in the current DAG. The USR-rnd variant allows the selection, with probability $1/|S|!$, of a backward permutation~$\tau$ that maximizes the acceptance probability \eqref{eq:a2}. Specifically, following the reasoning of Footnote~\ref{fn:ordering}, when $\tau$ matches some reverse topological ordering of $G$, the reduction of the pool of candidate parents of the nodes in the forward permutation~$\sigma$ is maximized, thereby minimizing $Z^\tau(G)$. We compare these variants both theoretically and empirically in Appendices~\ref{sec:comp} and \ref{sec:small}. The findings extend to the Markov blanket resampling (MBR) move, which also operates using a strongly positive set of proposal distributions and sequentially resamples parent sets.

\section{Experimental Comparisons}\label{sec:exp}

We have evaluated the performance of the DAG samplers on large, high-dimensional data sets to assess their robustness in challenging settings. Our study concerns the computational efficiency of \emph{different algorithms} for approximating the \emph{same posterior distribution}. The algorithms being agnostic to the specific statistical model employed (up to the modularity property), studying how well different models fit individual data sets or particular classes of data sets falls beyond the scope of the present work. We begin by describing the experimental setup, followed by the evaluation metrics and the corresponding results.

\subsection{Experimental Setup}

The next subsections introduce the compared samplers, the data sets used in the experiments, the Bayesian statistical model used, and the MCMC implementation.

\subsubsection{Algorithms}

We consider both single-move samplers and hybrid variants that combine multiple move types. For all moves, we employ the bottom-up pruning method by \citet{nikzad2025} described in Section \ref{sec:pru}. As discussed there, this method consistently bounds the total variation distance between the approximate and exact posteriors by $n\epsilon$, where $n$ is the number of nodes and $\epsilon$ is an input parameter set by the user. In our experiments, we set $\epsilon=2^{-10}$. Below, we list the moves considered in the experiments.

\begin{itemize}[leftmargin=1.5cm]
\item[\textbf{USR}] The unconstrained sequential resampling move, detailed in Section \ref{sec:bmh}. We compare only the USR-rev and USR-rnd variants, as we show in Appendices~\ref{sec:comp} and \ref{sec:small} that USR-id is inferior in performance. We set the block size $|S|$ to $\min\{\lfloor \sqrt{n} \,\rfloor, 10\}$; based on our preliminary studies (not shown), this rule of thumb works well for networks of different sizes, including the small networks examined in Appendix \ref{sec:small}.
\item[\textbf{EC}] The edge-constrained move, detailed in Section \ref{sec:ecpr}.
\item[\textbf{AC}] The ancestor-constrained move, detailed in Section \ref{sec:acpr}.
\item[\textbf{FBM}] The fast basic moves by \cite{nikzad2025}. 
\item[\textbf{REV}] The edge reversal move by \cite{Grzegorczyk2008}.
\item[\textbf{MBR}] The Markov blanket resampling move by \cite{Su2016}. We implement it within the duet sampling framework and, analogously to the USR move, we compare only the MBR-rev and MBR-rnd variants.  Appendices \ref{sec:comp} and \ref{sec:small} provide theoretical and empirical evidence that the original MBR-id variant performs poorly compared to these alternatives.
\item[\textbf{BG}] The blocked Gibbs move by \cite{Goudie2016}. We set the block size to 2, 3, or 4, for the move is computationally infeasible for larger values. For comparison, \cite{Goudie2016} set the maximum block size to 3.
\item[\textbf{MIX}] This is a hybrid move that combines the newly introduced moves: for every 12 USR-rev moves, it performs 4 USR-rnd moves, 3 EC moves, and 3 AC moves.
\item[\textbf{Gibby}] The Gibby algorithm by \cite{nikzad2025}. It uses a hybrid move that alternates between the FBM, REV, and MBR moves. We integrate the previously described MBR variants into the Gibby algorithm while keeping all other default parameters unchanged. We refer to the resulting variants as Gibby-\emph{rev} and Gibby-\emph{rnd}, depending on the MBR variant employed.
\end{itemize}

We have implemented the new moves and the BG move on top of the code base of Gibby \citep{nikzad2025}, while retaining the implementations of Gibby and its associated moves provided therein. Our version of the BG move aims at high performance following the architecture of \cite{Goudie2016} and extends it to support score pruning. Our code is available at \url{https://github.com/sums-of-products/duetto}. Further details of the implementations and computing environment are provided in Appendix~\ref{sec:env}.

\subsubsection{Data Sets}

We focus on large data sets consisting of 10\,000 data points sampled from large discrete benchmark Bayesian networks \citep{ScutariBNrepository}. These data sets induce posterior distributions that are challenging to explore and thereby suitable for studying the limitations of the samplers. Table~\ref{tab:dataset_summary_main} summarizes the benchmark networks considered in the experiments, reporting the number of nodes, the number of edges, and the \emph{maximum in-degree} (MID), i.e., the size of largest parent set in the network. We will refer to the generated data sets simply by the name of the generating network. For more details on the data sets, we refer the reader to Appendix~\ref{sec:env}.

\begin{table*}[tb]
\centering
{\small
\setlength{\tabcolsep}{6.0pt}
\begin{tabular}{lccc}
\toprule
Network & \#nodes & \#edges & MID \\
\midrule
%\BigCaps{Asia}   & 8   & 8   & 2 \\
%\BigCaps{Sachs}  & 11  & 17  & 3 \\
\BigCaps{Hailfinder} & 56  & 66  & 4 \\
\BigCaps{Andes}     & 223 & 338 & 6 \\
\BigCaps{Pigs}      & 441 & 592 & 2 \\
\bottomrule
\end{tabular}
}
\caption{Summary of the benchmark networks used in the experiments.}
\label{tab:dataset_summary_main}
\end{table*}

\subsubsection{Bayesian Model}

We used the BDeu score, with equivalent sample size equal to $1$, together with the so-called \emph{fair} structure prior \citep{Friedman2003,Eggeling2019}, which favors sparser graphs by setting the prior probability of $G$ to be proportional to $\prod_{i=1}^n \binom{n-1}{|P_i|}^{-1}$. To reduce the computational burden associated with parent-set scoring, the prior probability of parent sets whose size exceeds the MID of the data-generating network is set to zero. 

Furthermore, due to computational constraints, for the \BigCaps{Andes} data set we restricted the size of parent sets that contain nodes outside the \emph{top~$64$ candidate parents} to at most three. The top candidate parents are the nodes associated with the highest-scoring parent sets of size 1. This is the same approach as in \citet{nikzad2025}.

\subsubsection{MCMC Runs}

Ten independent runs per data set were carried out for each algorithm, each initialized from the empty DAG. Since the REV move requires that the current DAG contains at least one edge, the chains that perform only REV moves were initialized with 1\,000 FBM steps. Each algorithm was run for 15\,000 seconds. Table \ref{tab:nmoves} summarizes the total number of iterations completed by the single-move samplers. The move variants are grouped together, as their computational costs are approximately equal.

\begin{table*}[tb]
\centering
{\small
\setlength{\tabcolsep}{12.0pt}
\begin{tabular}{rlll}
\toprule
& \multicolumn{3}{c}{\text{\#steps}}\\
\cmidrule(lr){2-4}
Algorithm & \sc{Hailfinder} & \sc{Andes} & \sc{Pigs}\\
\midrule
\sc{USR}           & \num{2.6e8}& \num{1.2e7}    & \num{3.1e7} \\
\sc{EC \& AC}      & \num{7.7e8}    & \num{7.3e7}    & \num{1.1e8} \\[6pt]
\sc{FBM}           & \num{1.7e12} &  \num{1.7e12} &  \num{1.5e12} \\
\sc{REV}           &  \num{1.9e9} &  \num{6.1e7}  &  \num{6.2e8}  \\
\sc{MBR}           &  \num{1.6e9} &  \num{7.5e7}    &  \num{2.1e8} \\[6pt]
\sc{BG$\,(|S|=2)$} &  \num{1.5e8}             &  \num{2.7e6} &  \num{4.8e6} \\
\sc{BG$\,(|S|=3)$} &  \num{3.1e7}            &  \num{8.2e5}    &  \num{1.2e6} \\
\sc{BG$\,(|S|=4)$} &  \num{2.5e6}   &  \num{4.2e4}  &  \num{8.2e4}\\
\bottomrule
\end{tabular}
}
\caption{Number of Markov chain steps taken by the single-move algorithms in 15\,000 s.} %15\,000 seconds.}
\label{tab:nmoves}
\end{table*}

Varying thinning factors were applied to each algorithm to yield comparable sample sizes of DAGs for the evaluation statistics (see the next subsection). The thinning factors were also chosen to keep the memory requirement for storing the sampled DAGs manageable. After thinning, the sample sizes were roughly 100\,000 for \BigCaps{Hailfinder}, 30\,000 for \BigCaps{Andes}, and 20\,000 for \BigCaps{Pigs}.

\subsection{Results}

\begin{figure*}[!htbp]
\begin{center}
\setlength{\tabcolsep}{2pt}  % reduce horizontal padding
\small  % smaller font size

\begin{tabular}{>{\raggedright\arraybackslash}m{0.025\textwidth}
                >{\centering\arraybackslash}m{0.335\textwidth}
                >{\centering\arraybackslash}m{0.3185\textwidth}
                >{\centering\arraybackslash}m{0.323\textwidth}}

% -------------------- Column headers --------------------
&  \hailfinder{} ($n = 56$) &  \andes{} ($n = 223$) &  \pigs{} ($n = 441$) \\
\midrule
\addlinespace[12pt]

% ==================== Method 1 ====================
\rotatebox[origin=l]{90}{USR \& MIX} &
\includegraphics[width=\linewidth]{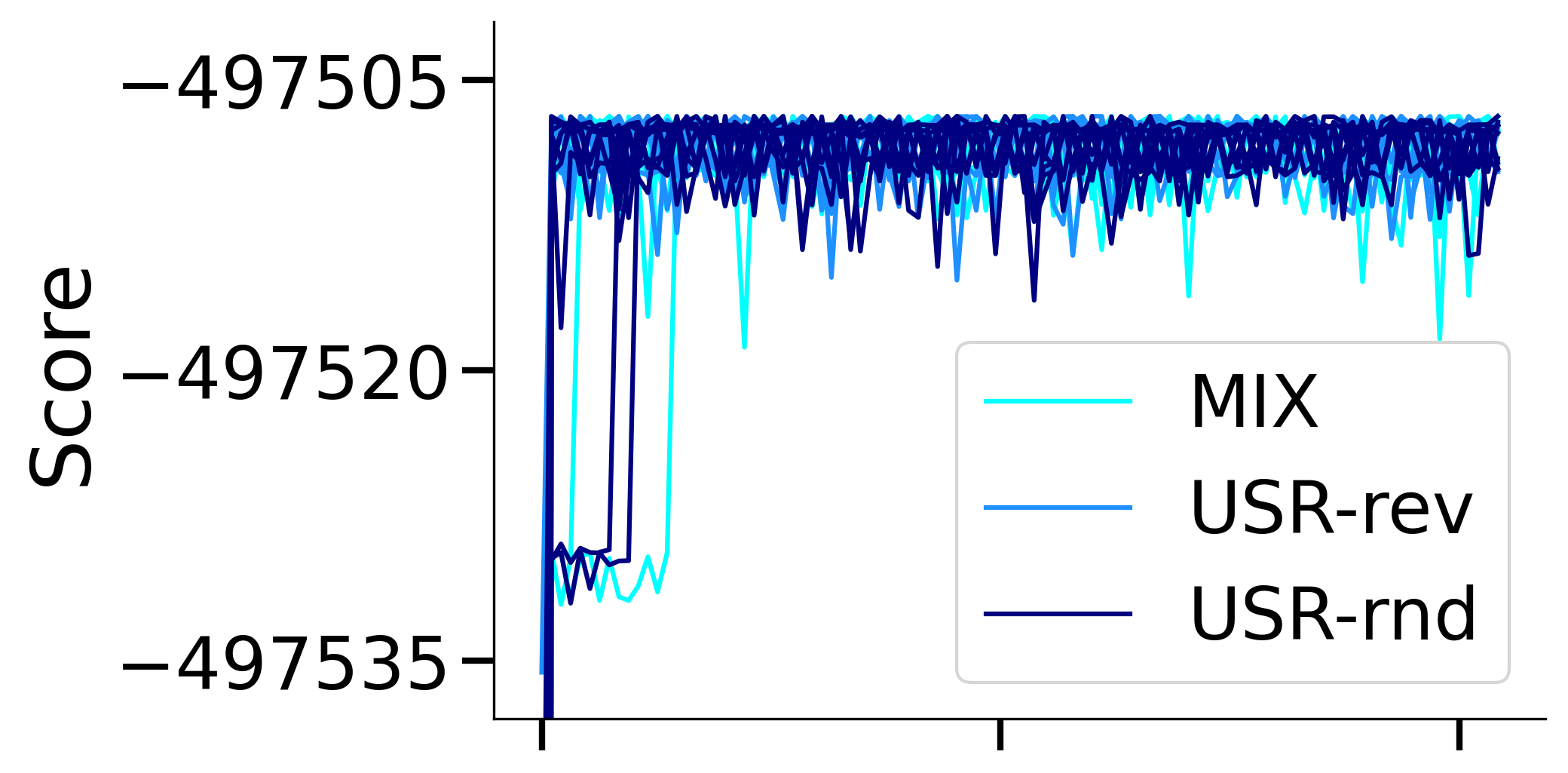} &
\includegraphics[width=\linewidth]{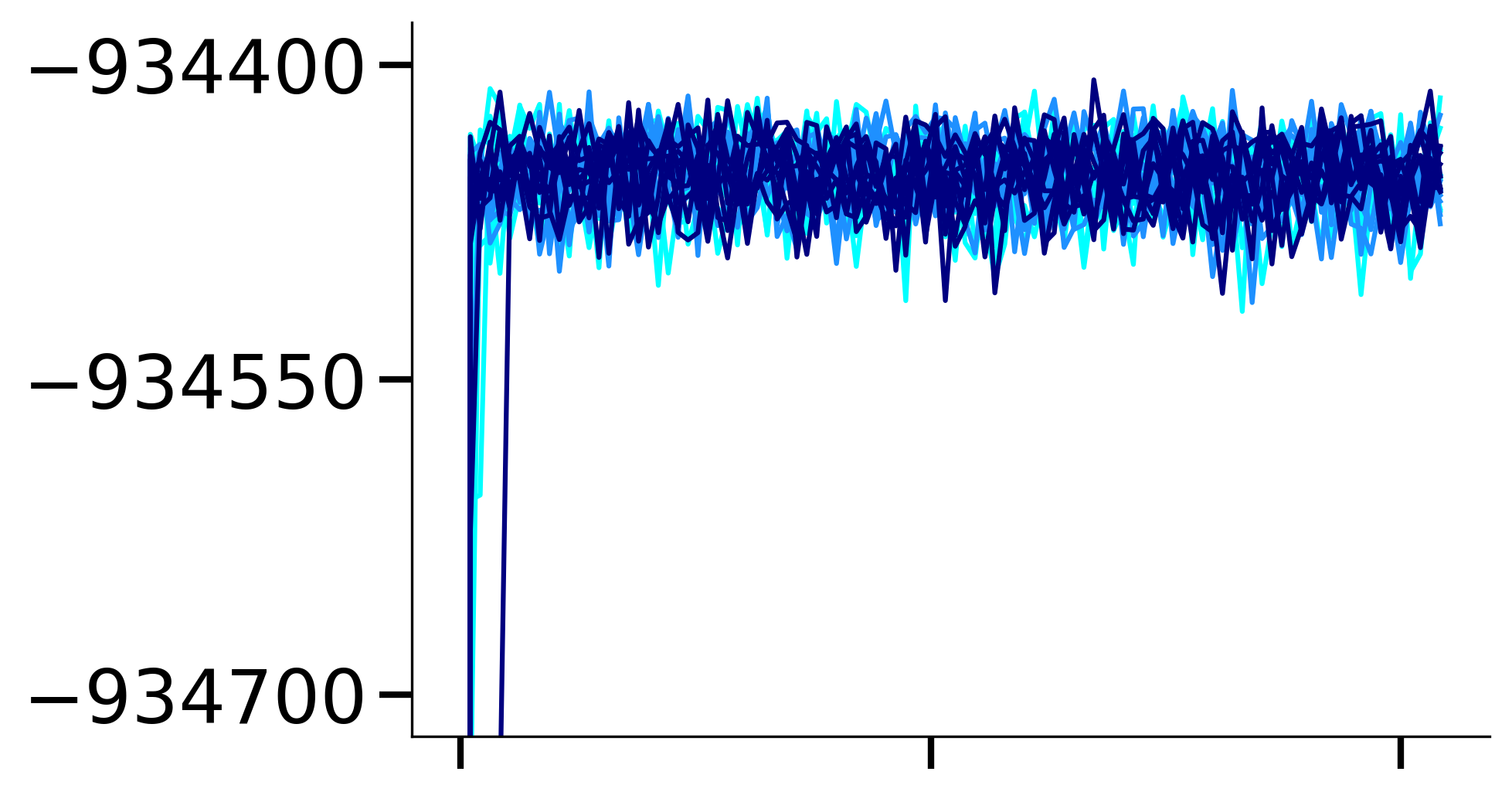} &
\includegraphics[width=\linewidth]{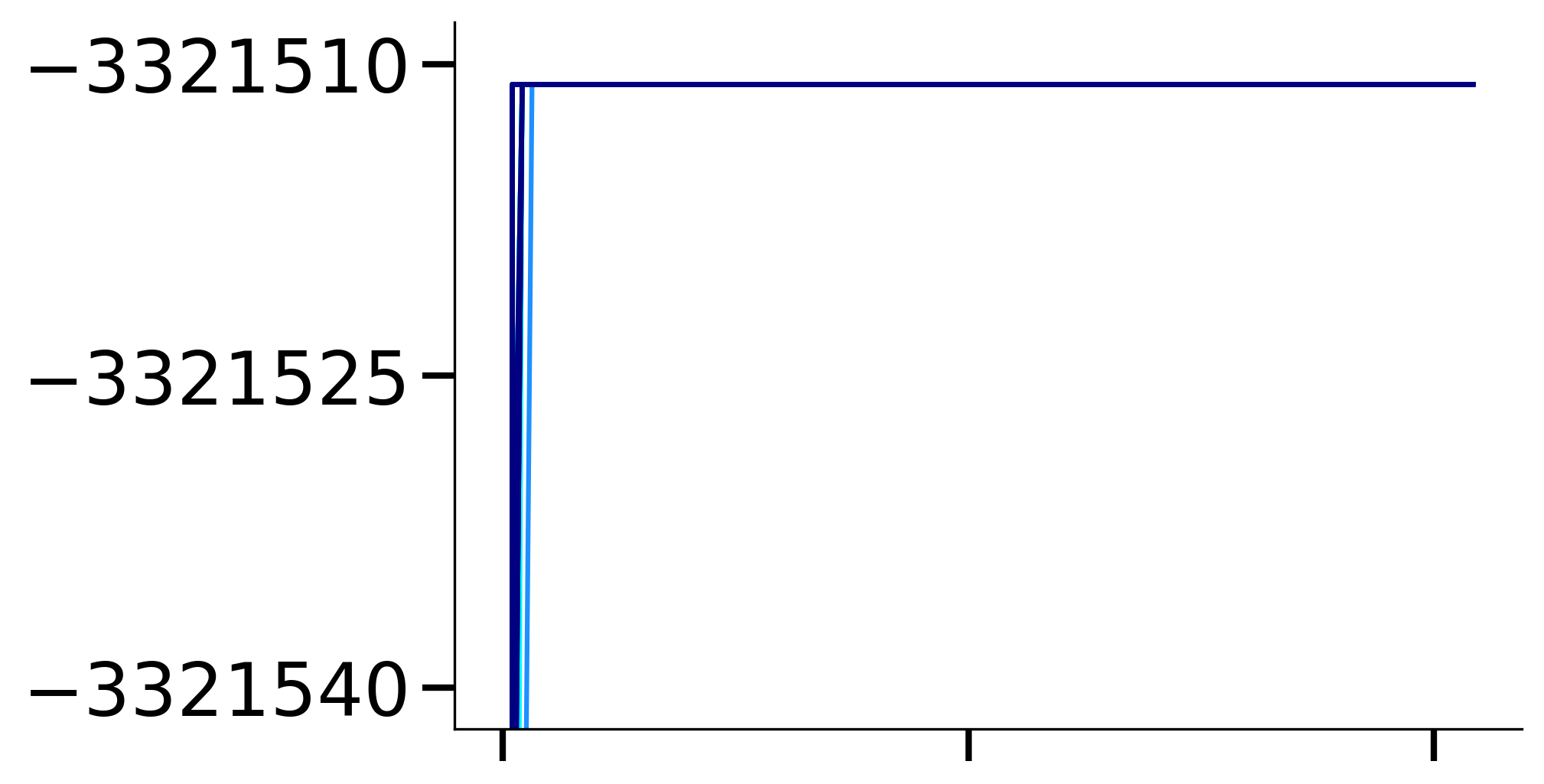} \\
\addlinespace[6pt]

% ==================== Method 2 ====================
\rotatebox[origin=l]{90}{EC \& AC} &
\includegraphics[width=\linewidth]{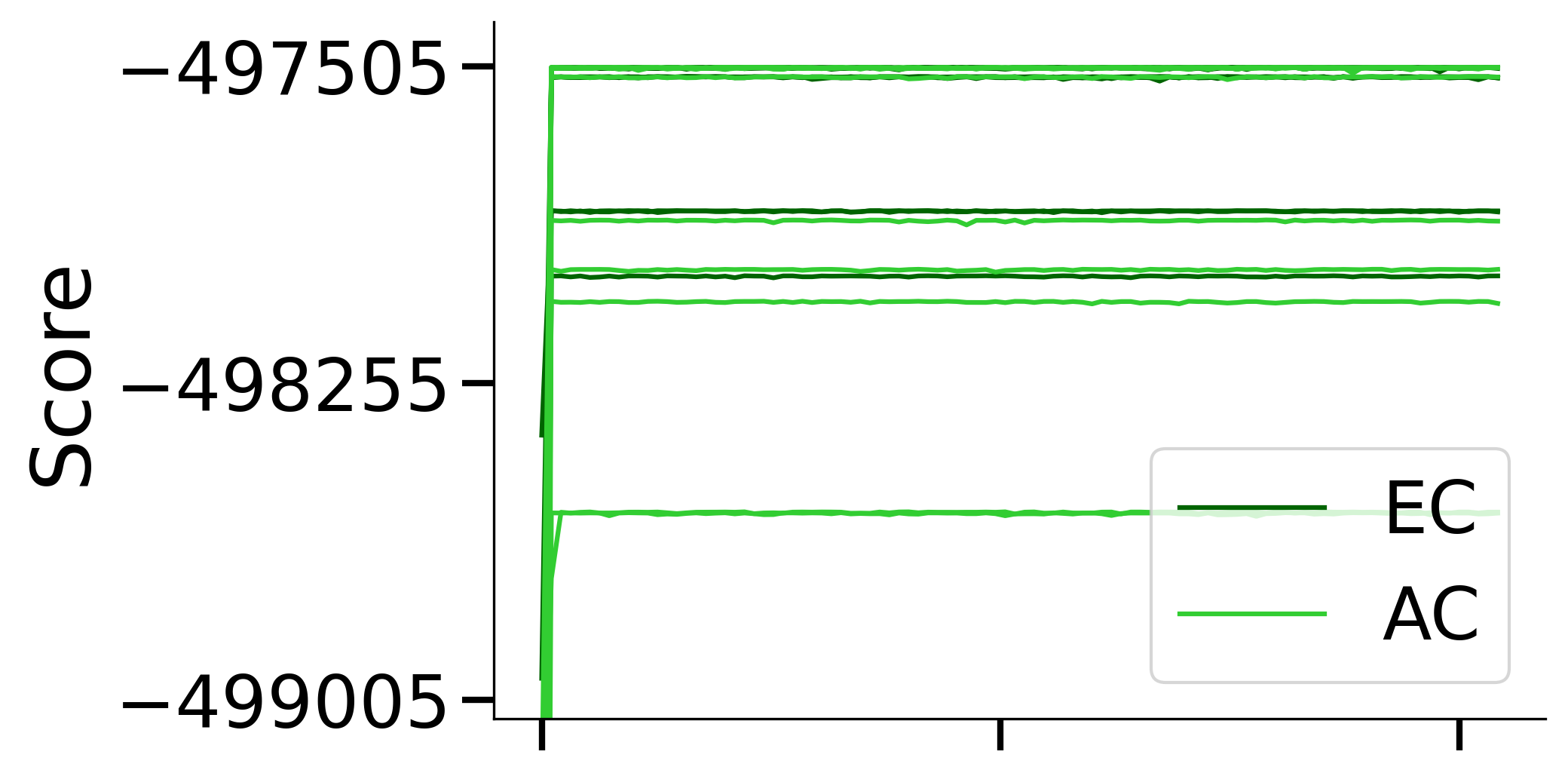} &
\includegraphics[width=\linewidth]{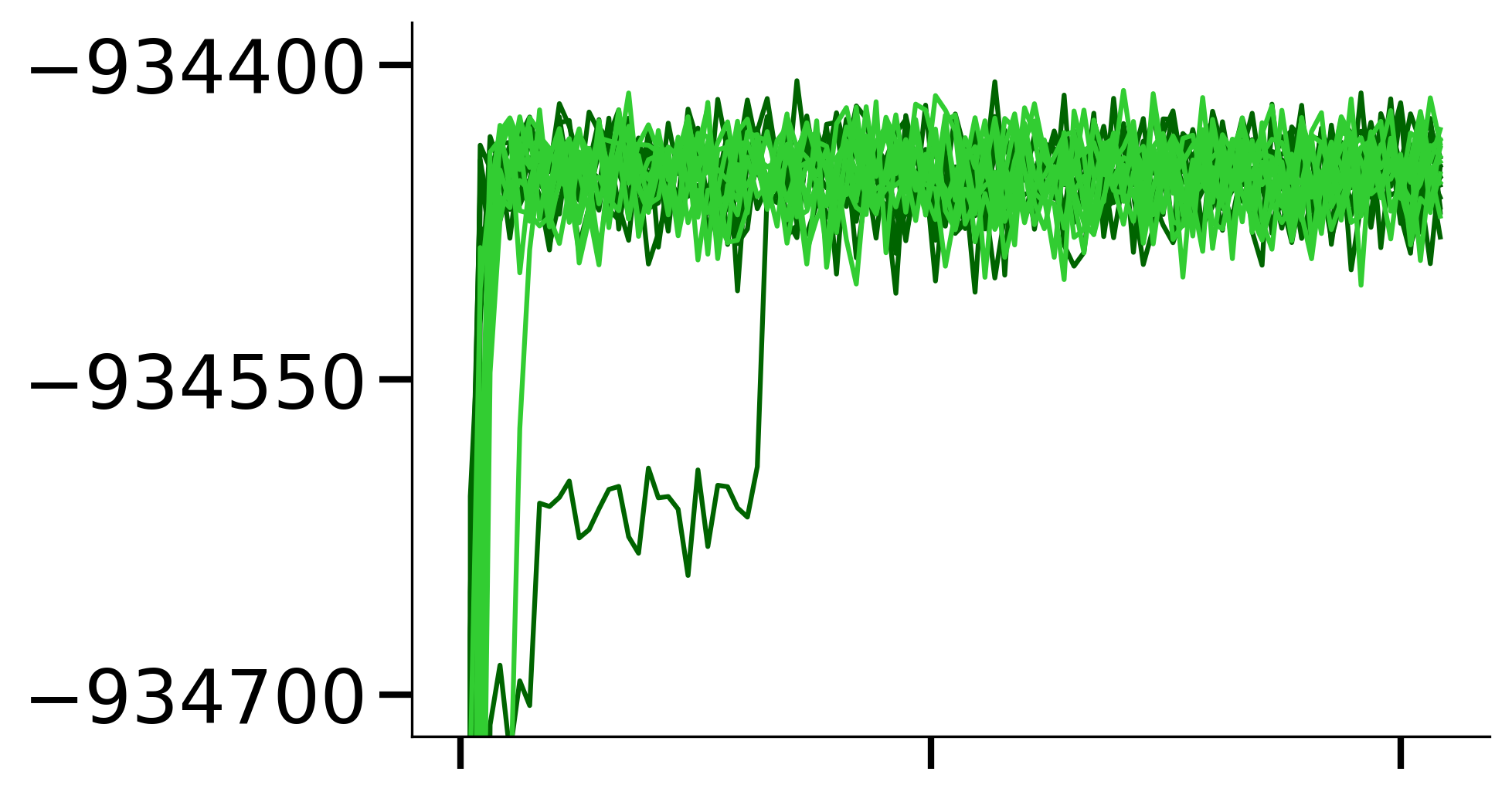} &
\includegraphics[width=\linewidth]{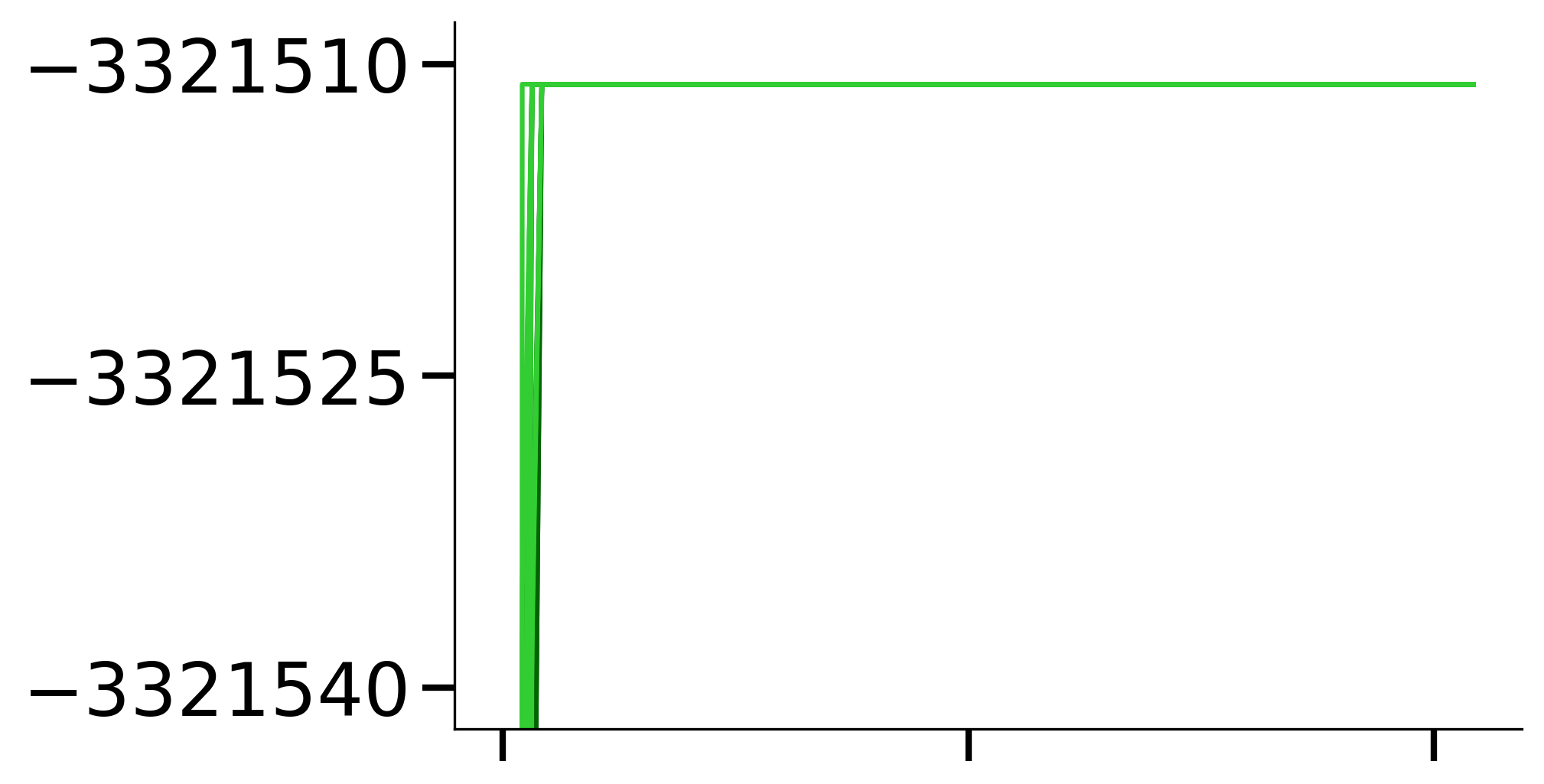} \\
\addlinespace[6pt]

% ==================== Method 3 ====================
\rotatebox[origin=l]{90}{BG}&
\includegraphics[width=\linewidth]{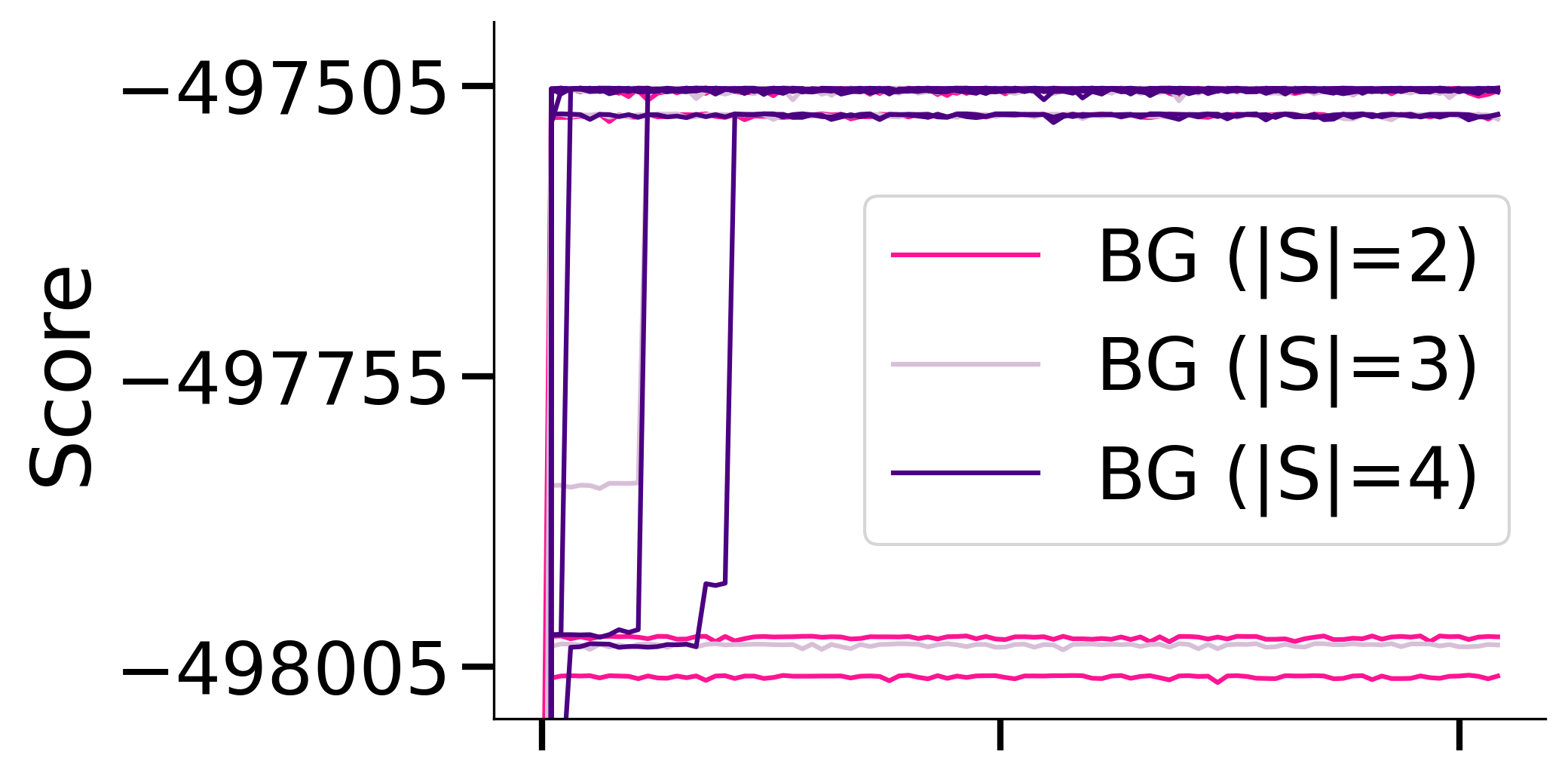} &
\includegraphics[width=\linewidth]{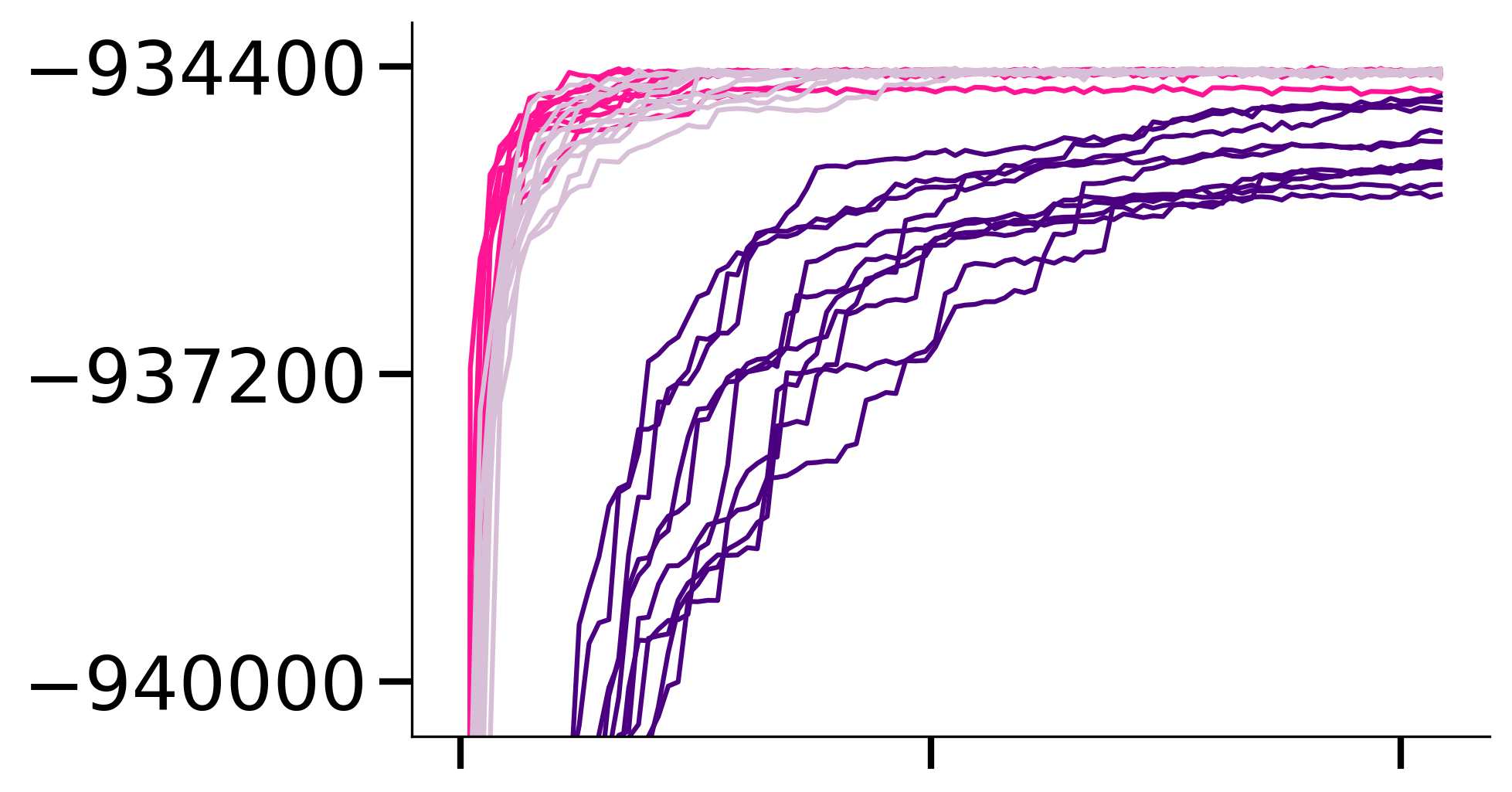} &
\includegraphics[width=\linewidth]{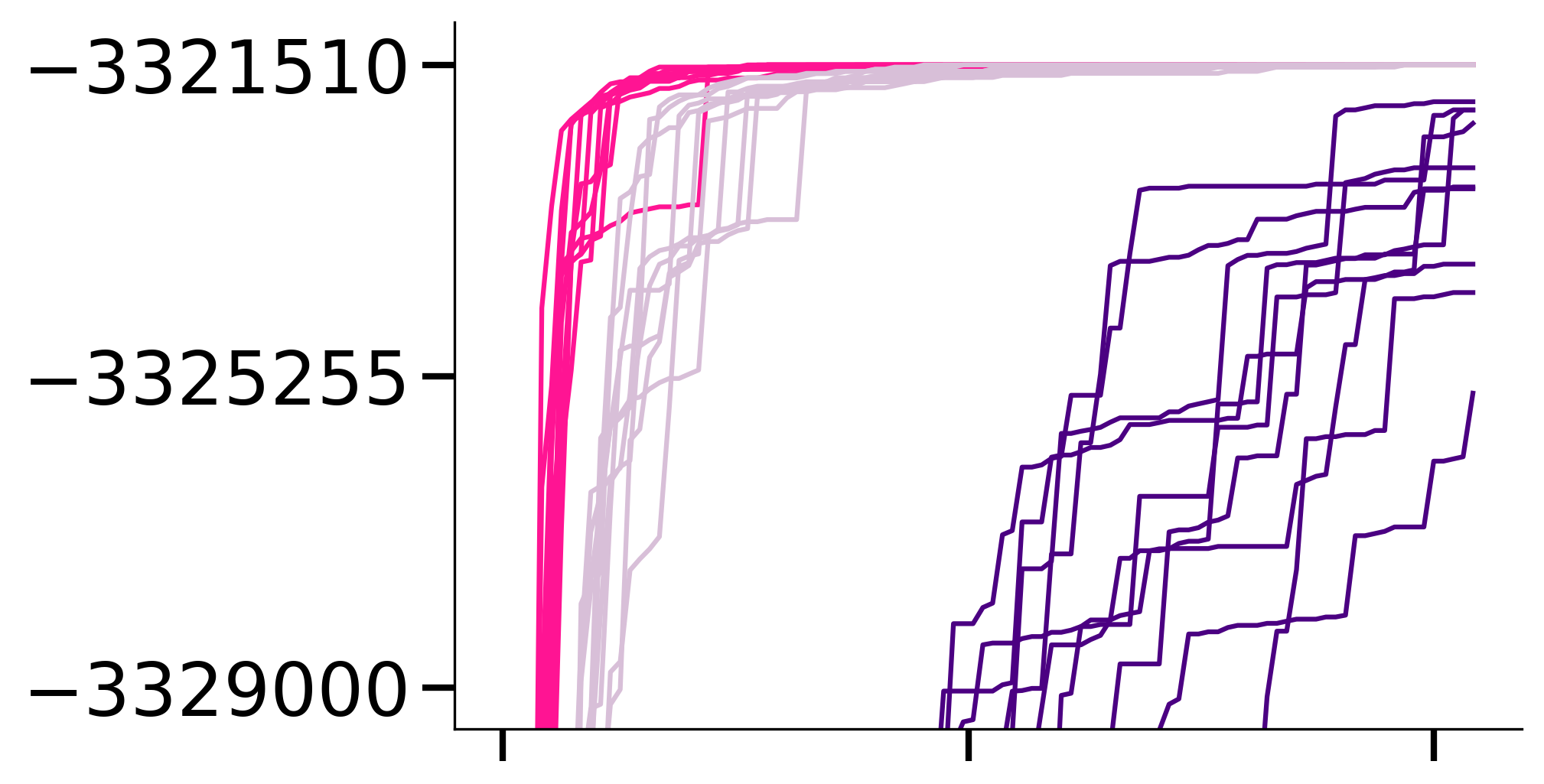} \\
\addlinespace[6pt]

% ==================== Method 4 ====================
\rotatebox[origin=l]{90}{REV \& FBM}&
\includegraphics[width=\linewidth]{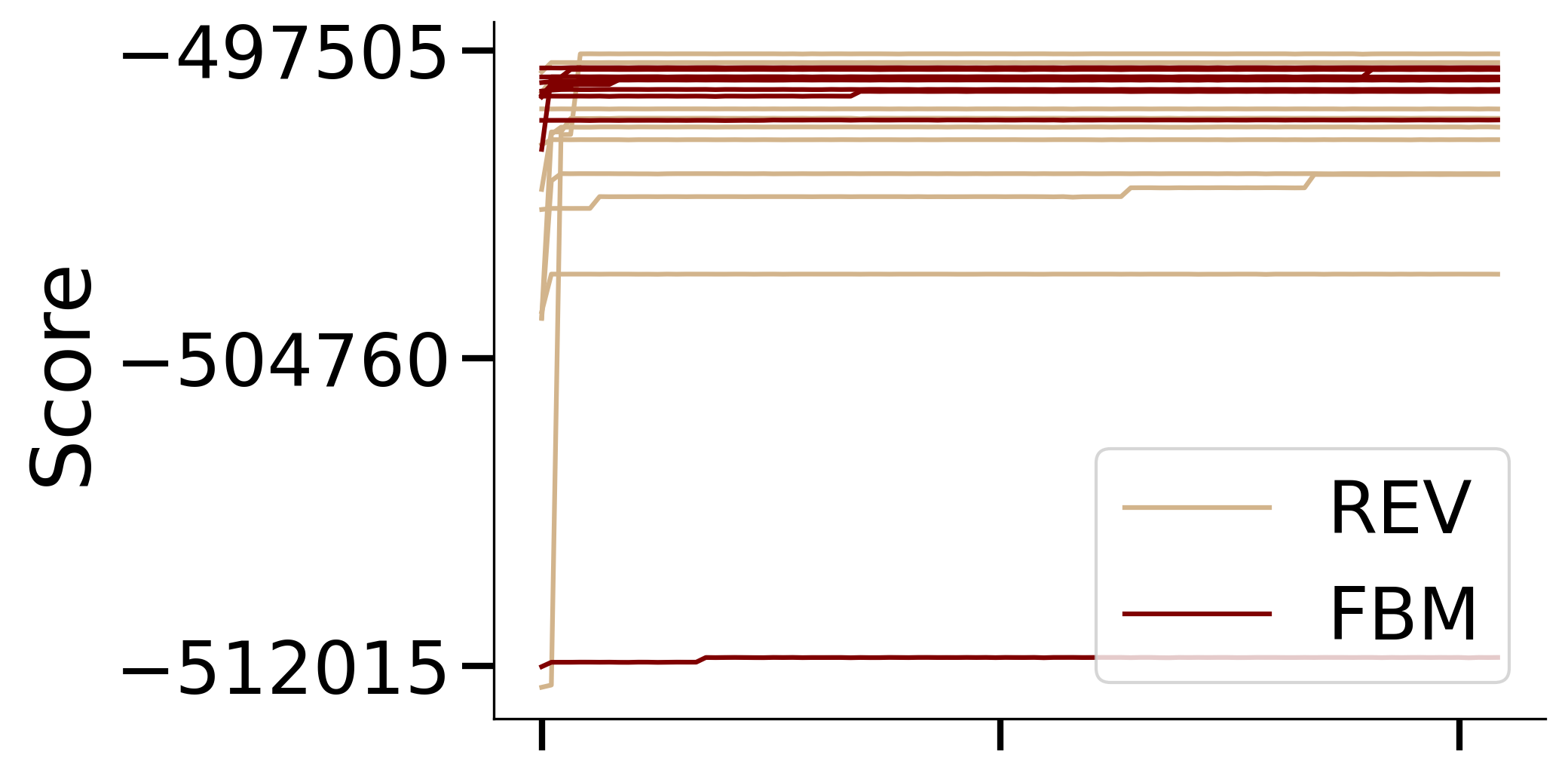} &
\includegraphics[width=\linewidth]{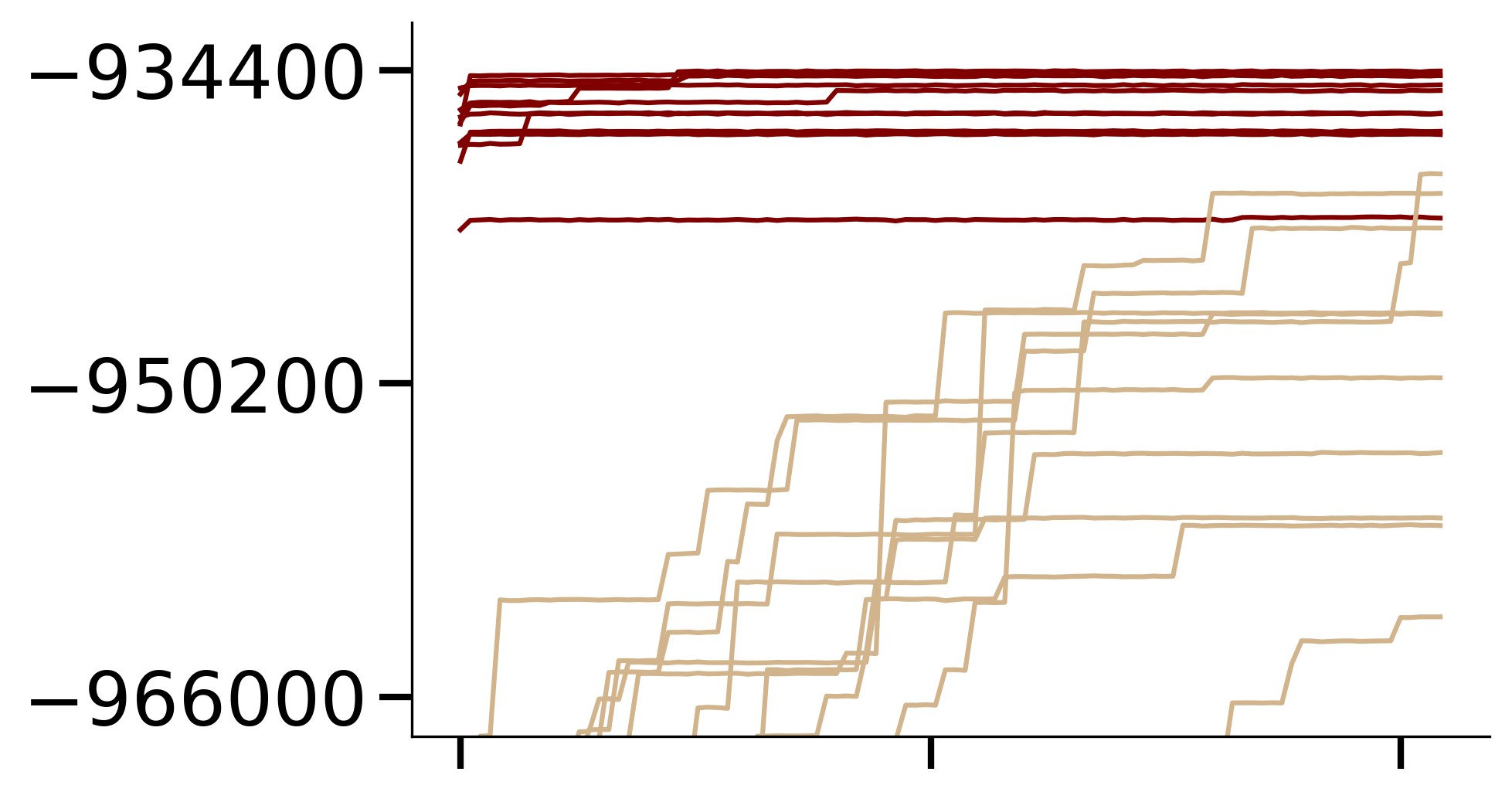} &
\includegraphics[width=\linewidth]{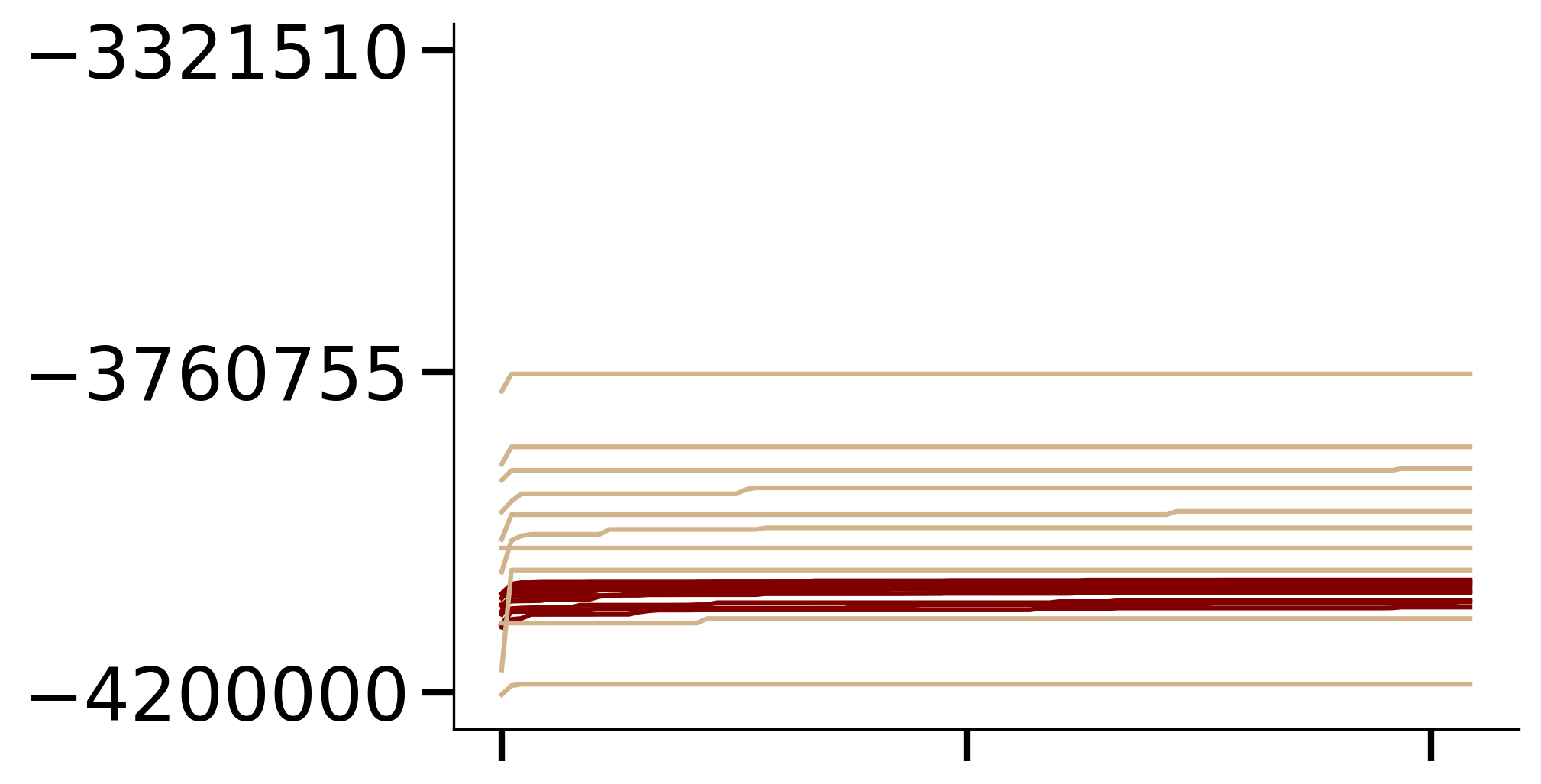} \\
\addlinespace[6pt] 

% ==================== Method 5 ====================
\rotatebox[origin=l]{90}{MBR}&
\includegraphics[width=\linewidth]{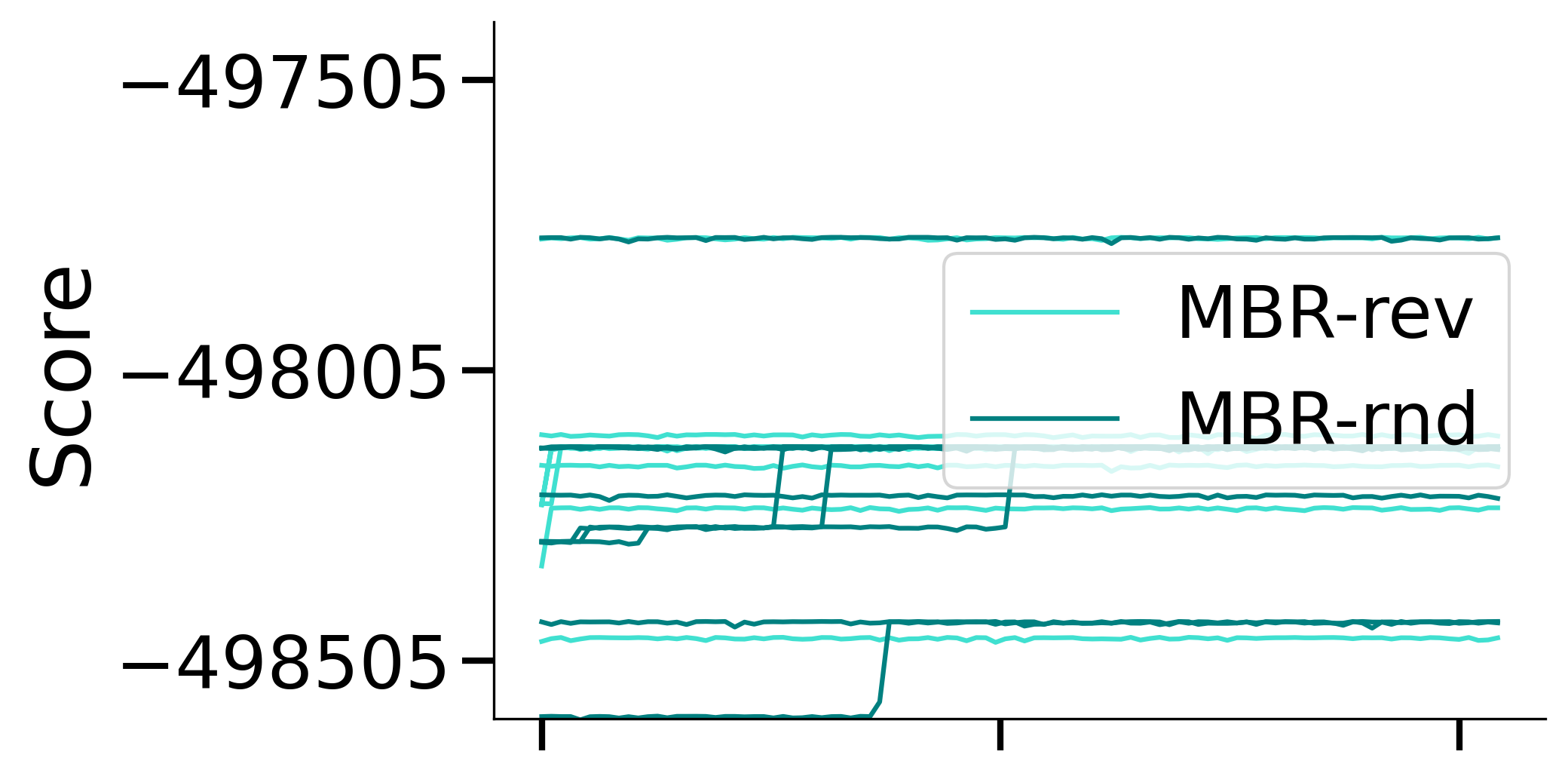} &
\includegraphics[width=\linewidth]{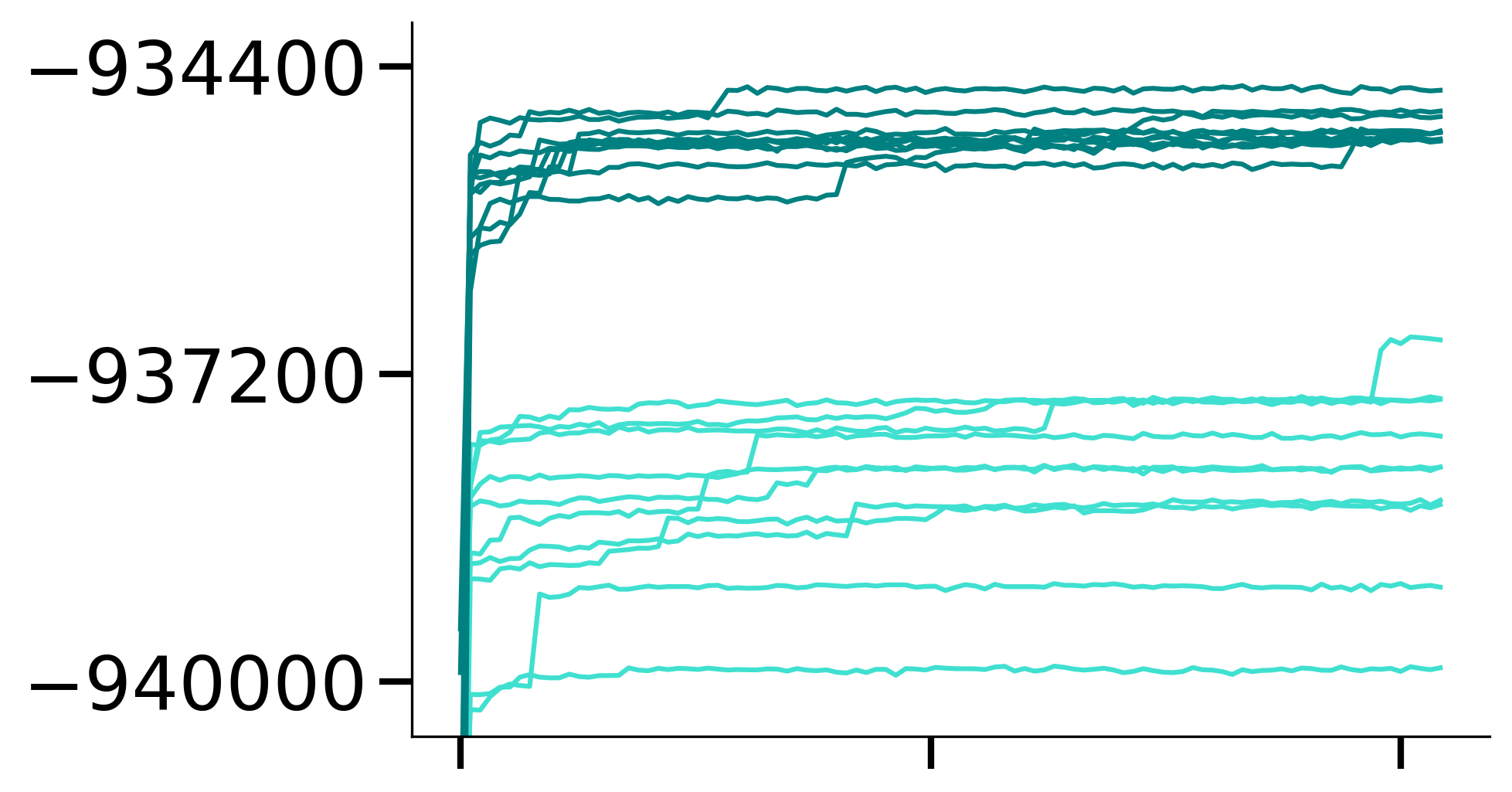} &
\includegraphics[width=\linewidth]{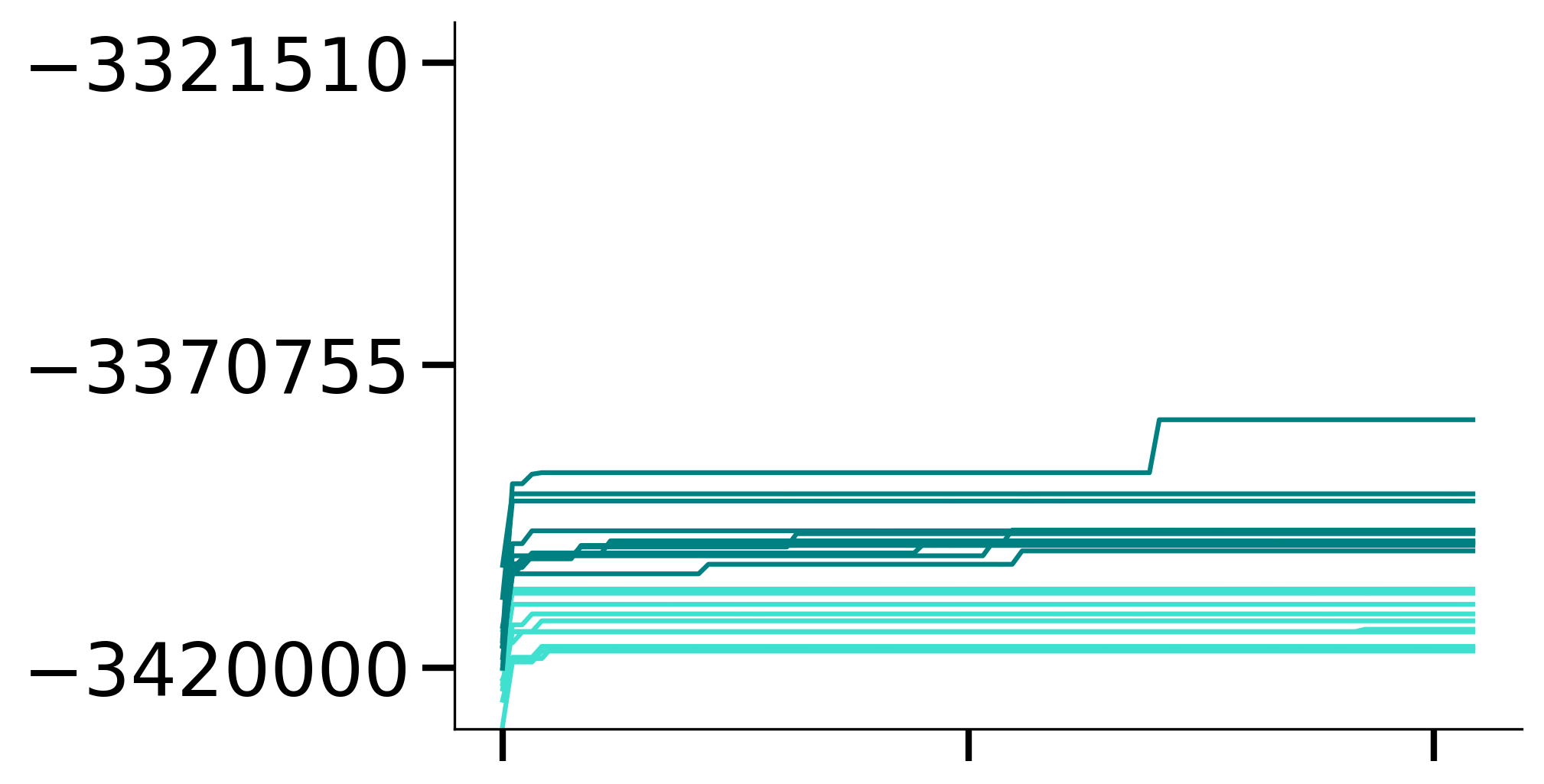} \\
\addlinespace[6pt]

% ==================== Method 6 ====================
\rotatebox[origin=l]{90}{Gibby}&
\includegraphics[width=\linewidth]{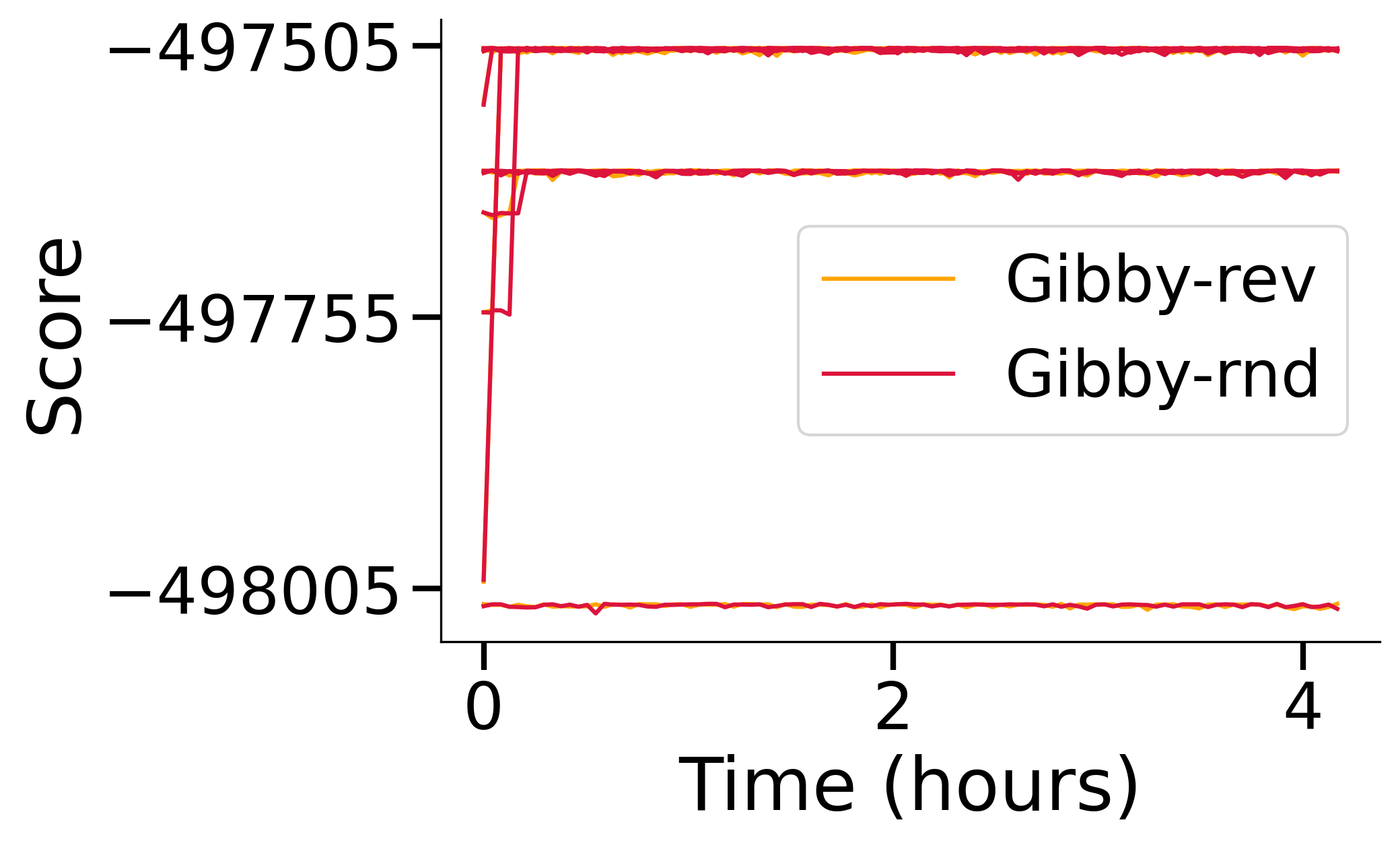} &
\includegraphics[width=\linewidth]{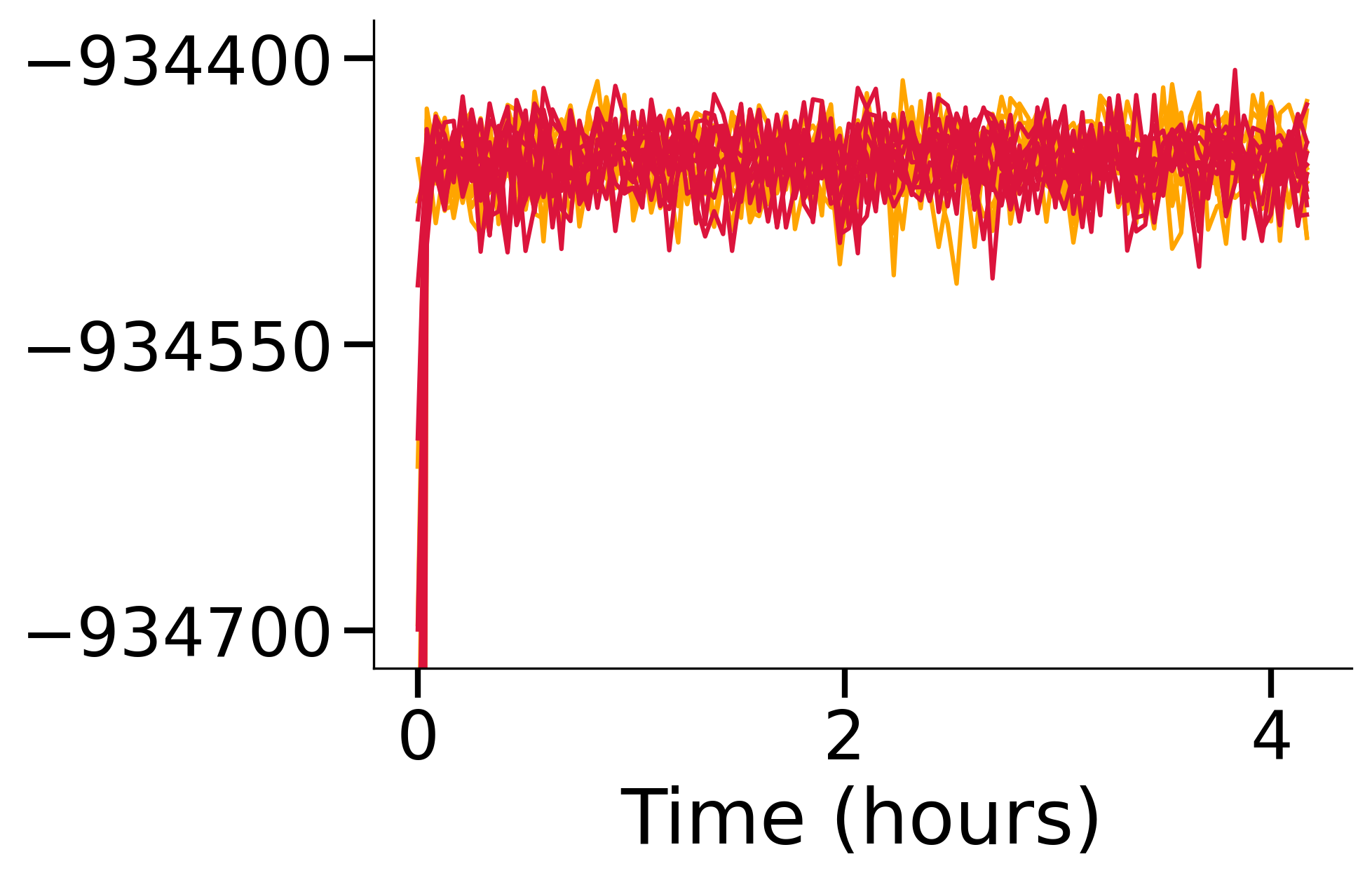} &
\includegraphics[width=\linewidth]{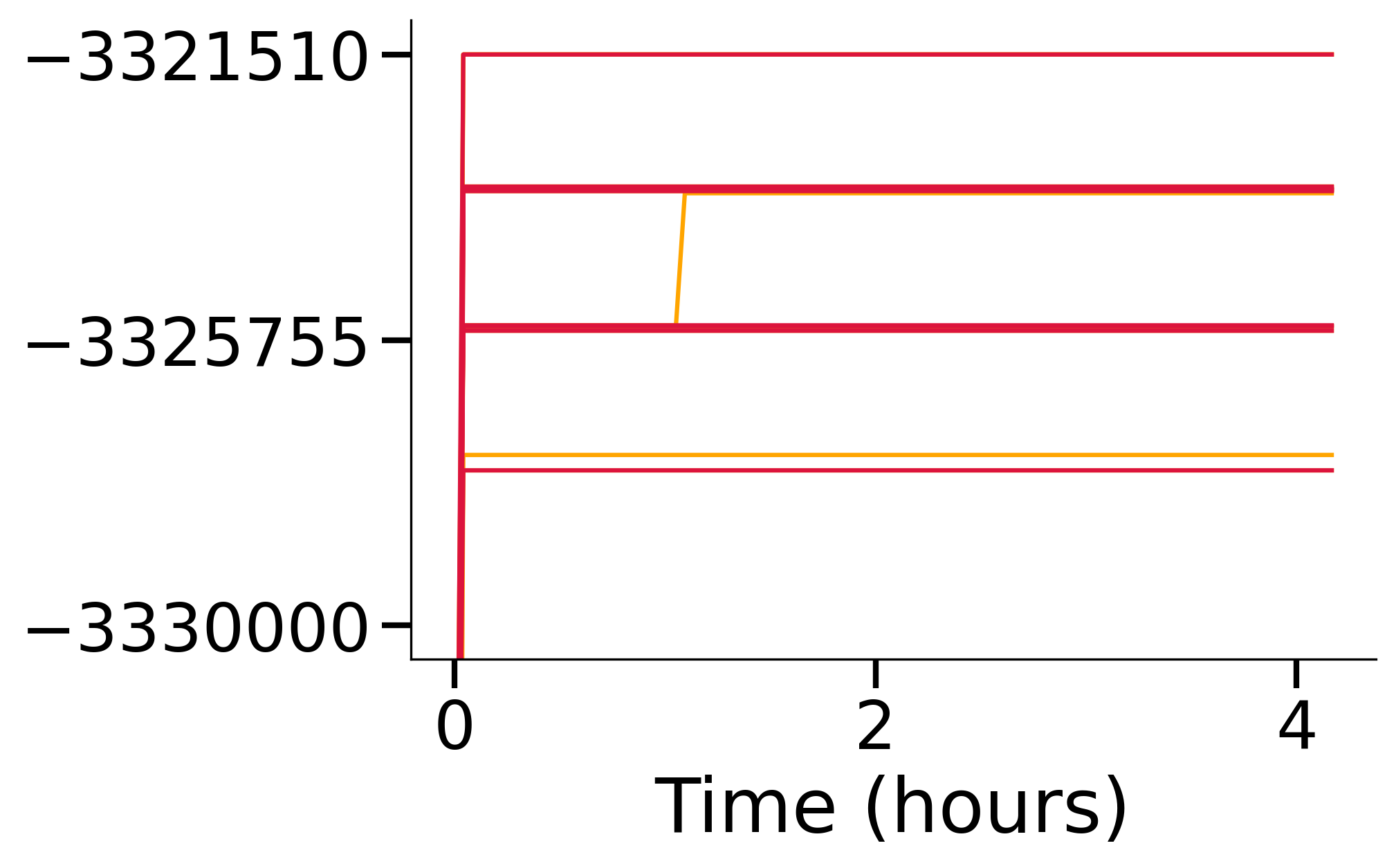} \\

\end{tabular}

\normalsize
\caption{Traceplots of MCMC runs for data sets sampled from large benchmark networks. Columns correspond to data sets. Rows correspond to (groups of) algorithms. For each algorithm, ten independent runs are displayed. For each run, the log-unnormalized posterior probability (log-score) is shown as a function of running time for 100 evenly spaced DAG samples.} 
%The x-axis represents the sampling time; the y-axis shows the log-unnormalized posterior probability. For each run, 100 sampled DAGs are visualized. 
\label{fig:large}
\end{center}
\end{figure*}

By visually inspecting the traceplots (Figure~\ref{fig:large}), we can identify cases where an algorithm fails to consistently converge close to the posterior distribution. FBM, REV, and MBR fail across all data sets and move variants, while Gibby performs well only on \BigCaps{Andes}. For BG, the traceplots indicate two different problems. On \BigCaps{Hailfinder} individual runs can become trapped near local modes regardless of block size. On \BigCaps{Andes} and \BigCaps{Pigs}, it is unclear whether exploration is simply slow or whether the runs become trapped near local modes. Overall, BG shows relatively good performance only on \BigCaps{Andes} with block size 2 and on \BigCaps{Pigs} with block sizes 2 and 3. In contrast, the newly proposed moves consistently reach the main mode of the posterior distribution rapidly, with the only exceptions being EC and AC on \BigCaps{Hailfinder}.

As traceplots are too coarse to reveal more subtle issues in mixing and convergence, we also inspect two additional convergence diagnostics (Table~\ref{table:experiments}). First, for each data set, we consider the highest logarithmic score $\ell_A^a$ (i.e., unnormalized log-posterior probability) of the DAGs visited by the $a$-th run of algorithm $A$. The highest log-scores provide us insight into the algorithm's ability to climb high-probability regions and escape local modes. We record both the maximum~$\ell_A^\text{max} := \max_a \ell_A^a$ and the minimum~$\ell_A^\text{min} := \min_a \ell_A^a$ log-score over the ten runs. Denoting by $\ell_*$ the largest $\ell_A^\text{max}$ over all algorithms $A$, we report the deviations
\bes
	\Delta_A^\text{min} := \ell_* - \ell_A^\text{max}
	\quad \textrm{and} \quad 
	\Delta_A^\text{max} := \ell_* - \ell_A^\text{min}\,.	
\ees
Unlike the raw $\ell$-values, the $\Delta$-values are comparable across data sets and enable displaying simple numbers. Specifically, if $\Delta_A^\text{max}$ equals $0$ for a data set, then algorithm $A$ found the highest scoring DAGs consistently in all ten runs.

We also report the \emph{worst edge probability difference} (WEPD), defined as the largest absolute difference in the edge probability estimates between any two independent runs, computed over all runs of a given algorithm. More precisely,
\bes
	\mathrm{WEPD}
	= \max_{a,b}\, \max_{i,j}\, | \hat{p}_{ij}^{a} - \hat{p}_{ij}^{b} |\,,
\ees
where $\hat{p}_{ij}^{a}$ is the estimated posterior probability of the presence of edge $ij$ based on the $a$-th MCMC run. The estimate is simply the relative frequency 
\bes
	\hat{p}_{ij}^{a} = \frac{1}{T} \sum_{t=1}^T \mathbf{1}\{ i j \text{ is an edge in } G^{a, t} \}\,,   
\ees
where $G^{a, t}$ is the $t$-th sampled DAG in run $a$, and $T$ is the total number of sampled DAGs in the chain.
The WEPD aims to assess whether independent runs converge to the same distribution by comparing the empirical marginal distributions on node pairs (potential edges). While measuring convergence to the exact distributions on large networks is computationally infeasible, in Appendix~\ref{sec:small} we present results against the exact marginal posterior distributions on small networks.

\begin{table*}[tb]
\centering
{\small
\setlength{\tabcolsep}{6.0pt}  % default is 6pt
\begin{tabular}{rccS[table-format=1.4] ccS[table-format=1.3] ccS[table-format=1.3]}
\toprule
& \multicolumn{3}{c}{\sc{Hailfinder}}
& \multicolumn{3}{c}{\sc{Andes}}
& \multicolumn{3}{c}{\sc{Pigs}}\\
& \multicolumn{3}{c}{$\ell_* = -497\,506.9$}
& \multicolumn{3}{c}{$\ell_* = -934\,406.0$}
& \multicolumn{3}{c}{$\ell_* = -3\,321\,511.0$}\\
\cmidrule(lr){2-4}
\cmidrule(lr){5-7}
\cmidrule(lr){8-10}
Algorithm 
& $\Delta^\text{min}$ & $\Delta^\text{max}$ & {$\mathrm{WEPD}$}
& $\Delta^\text{min}$ & $\Delta^\text{max}$ & {$\mathrm{WEPD}$}
& $\Delta^\text{min}$ & $\Delta^\text{max}$ & {$\mathrm{WEPD}$}\\
\midrule
USR-rev       & \textbf{0} & \textbf{0} & \textbf{0.0074} & \textbf{0} & \textbf{0} & 0.027          & \textbf{0} & \textbf{0} & 0.025\\
USR-rnd       & \textbf{0} & \textbf{0} & 0.095           & \textbf{0} & \textbf{0} & 0.044          & \textbf{0} & \textbf{0} & \textbf{0.018}\\
EC            & \textbf{0} & \num{5e2}  & 1.00            & \textbf{0} & \textbf{0} & 0.29           & \textbf{0} & \textbf{0} & 0.034\\
AC            & \textbf{0} & \num{1e3}  & 1.00            & \textbf{0} & \textbf{0} & 0.080          & \textbf{0} & \textbf{0} & 0.032\\
MIX           & \textbf{0} & \textbf{0} & 0.13            & \textbf{0} & \textbf{0} & 0.041          & \textbf{0} & \textbf{0} & 0.019\\[6pt]
FBM           & \num{4e2}  & \num{1e5}  & 1.00            & \textbf{0} & \num{7e3}  & 1.00           & \num{7e5}  & \num{8e5}  & 1.00\\
REV           & \num{7e1}  & \num{5e3}  & 1.00            & \num{5e3} & \num{2e4}  & 1.00           & \num{4e5}  & \num{9e5}  & 1.00\\
MBR-rev       & \num{3e2}  & \num{6e2}  & 1.00            & \num{3e3}  & \num{5e3}  & 1.00           & \num{6e4}  & \num{8e4}  & 1.00\\
MBR-rnd       & \num{3e2}  & \num{9e2}  & 1.00            & \num{2e2}  & \num{6e2}  & 1.00           & \num{6e4}  & \num{8e4}  & 1.00\\
Gibby-rev     & \textbf{0} & \num{5e2}  & 1.00            & \textbf{0} & \num{1e0}  & \textbf{0.015} & \textbf{0} & \num{6e3}  & 1.00\\
Gibby-rnd     & \textbf{0} & \num{5e2}  & 1.00            & \textbf{0} & \textbf{0} & 0.018          & \textbf{0} & \num{6e3}  & 1.00\\[6pt]
BG$\,(|S|=2)$ & \textbf{\num{0}}  & \num{1e3}  & 1.00            &  \textbf{0}     & \num{1e2}     & 1.00        & \textbf{0}  & \textbf{0}  & 0.65\\
BG$\,(|S|=3)$ & \textbf{\num{0}}  & \num{5e2}  & 1.00            & \textbf{0}     & 1   & 0.48         & \textbf{0}  & \textbf{0} & 0.88\\
BG$\,(|S|=4)$ & \textbf{\num{0}}  & \num{2e1}  & 1.00            & \num{2e2}      & \num{1e3}     & 1.00        & \num{4e2}  & \num{4e3}  & 1.00\\
\bottomrule
%\multicolumn{5}{l}{\scriptsize $^{\dagger}$a} &
%\multicolumn{5}{l}{\scriptsize $^{\ddagger}$b}
\end{tabular}
}
\caption{Comparison of the highest scores found and the differences of edge probability estimates across 10 independent MCMC runs. Smaller values are better.
}
\label{table:experiments}
\end{table*}

These metrics, displayed in Table~\ref{table:experiments}, suggest that consistently reaching the highest-probability mode across the ten runs is generally associated with consistent edge-probability estimates across runs. There are, however, notable exceptions. For BG on {\sc Andes} with block size 2 and on {\sc Pigs} with block sizes 2 and 3, we have $\Delta^\text{max}=0$, while the corresponding $\mathrm{WEPD}$ values are $0.48$, $0.65$, and $0.88$, respectively, which are relatively large. Indeed, the traceplots indicate that BG converges slowly, with several runs reaching the highest-probability mode only after hours of exploration. A similar discrepancy occurs for EC on {\sc Andes}: although $\Delta^\text{max}=0$, the $\mathrm{WEPD}$ is 0.29. This can be explained by a single run that spent a relatively long initial period in lower-probability regions before reaching the highest-probability modes, as is clearly visible in the traceplot. On the other hand, Gibby-rev on {\sc Andes} shows that a low $\mathrm{WEPD}$ can be achieved even when the worst run does not reach the highest-probability mode ($\Delta^\text{max} \approx 1$). This should not be interpreted as a failure case. Rather, it indicates that the posterior contains far fewer DAGs attaining the highest probability than DAGs with slightly lower probability. Consequently, DAGs in the highest mode might not get sampled by the Markov chain, without this affecting the estimation of low-dimensional marginals.

{\sc Hailfinder} appears to be the most challenging of the three benchmarks despite its relatively small number of nodes ($n = 56$): only three algorithms achieve a small $\mathrm{WEPD}$. On that data set, the best-performing algorithm is USR-rev by large $\mathrm{WEPD}$-margins to the second-best USR-rnd and the third-best MIX; for the latter two, the larger $\mathrm{WEDP}$ values are explained by a few unlucky runs that required longer ``burn-in'' periods (see the traceplots). 

\subsection{Conclusions}

In summary, the results reflect the limitations of the FBM, REV, MBR, and BG moves discussed in Sections \ref{sec:pmt} and \ref{sec:eccomp}, and are well aligned with the motivations underlying the newly proposed USR, EC, and AC moves. However, the EC and AC moves can fail when used alone (cf.\ the results on {\sc {Hailfinder}}). This is because they orphan and resample the parent sets of only two nodes, which in some instances can hinder transitions to high-probability DAGs. The {\sc Hailfinder} traceplots reveal similar limitations for the BG move, which is only feasible with small block sizes. Furthermore, when the block size is 4, BG fails to reach high-probability regions of the DAG posterior within the allotted time in the {\sc {Andes}} and {\sc {Pigs}} experiments. The USR move variants are more successful because they are fast while still being able to orphan and informatively resample the parent sets of a larger number of nodes in a single iteration.

\section{Concluding Remarks}\label{sec:cr}

Structure MCMC currently stands out as the most promising method for computing summaries of posterior distributions of DAGs \citep{nikzad2025}. The main contributions of the present work, namely duet sampling and the new moves in the DAG space, further advance the framework, enabling reliable Bayesian inference on large data sets. Below, we discuss the key implications, highlight some remaining challenges, and outline directions for future research.

The key advantage of duet sampling, compared to basic Metropolis--Hastings (MH), is the possibility to tune the acceptance rule by balancing the ``forward proposal distribution'' by a possibly different ``backward proposal distribution'', yielding a more informative acceptance criterion. While duet sampling can be viewed as a special case of even more general algorithms for constructing reversible Markov chains, specifically the involutive MCMC by \cite{Neklyudov2020} and the related earlier framework by \cite{storvik2011}, duet sampling gives a novel concrete template that we may expect to find applications also in other settings. 

The new moves in the DAG space, namely, the edge-constrained (EC) move, the ancestor-constrained (AC) move, and the unconstrained sequential sampling (USR) move, all leverage duet sampling. For the EC and AC moves, the duet sampling formulation is a way to show that the chain is reversible with the desired stationary distribution.  
For the USR move, basic MH corresponds to one possible move variant, which, however, appeared to perform poorly: the basic MH acceptance ratio favors the preservation of the ancestor relationships, discouraging the acceptance of DAGs that reverse some ancestor relationships of the current DAG. In contrast, duet sampling enables the construction of move variants with opposite behavior, encouraging the acceptance of DAGs that differ substantially from the current DAG. 

In our empirical study, the USR-rev move, in which the backward proposal distribution reverses the permutation of the selected node block, achieved the best overall performance among the individual moves across the benchmark data sets and repeated runs. Although we did not encounter any failure cases for USR-rev in our experiments, its performance is not guaranteed across all data sets. Combining different moves can mitigate the limitations of individual moves and promote more reliable exploration. For this reason, we recommend MIX, which combines the USR-rev, USR-rnd, EC, and AC moves, as the most robust choice. The EC and AC moves complement the USR move by rapidly identifying small high-probability substructures, while the USR move enables flexible jumps between high-probability DAGs that differ in several parent sets.

One major challenge of the taken approach is the memory and time requirements of precomputing the needed local scores. When the the number of nodes is large and nodes are allowed to have many parents, even if score puning is effective, we may have to restrict the possible parents of each node to some pre-selected set of candidate parents, a common practice also in previous works \citep{Friedman2003,nikzad2025,Viinikka2020}. For example, for the {\sc Andes} data set, with 223 nodes, we employed a simple heuristic to select 64 candidate parents for each node. While we may expect the selected candidates contain the most ``relevant'' nodes, we currently have no good way to bound the bias caused by ignoring the remaining possible parents. In fact, the study of \cite{Viinikka2020} shows that various natural heuristics for selecting candidate parents can miss some important nodes in the candidates. These issues warrant further investigations. 

% Acknowledgements and Disclosure of Funding should go at the end, before appendices and references

\acks{
%All acknowledgements go at the end of the paper before appendices and references. Moreover, you are required to declare funding (financial activities supporting the submitted work) and competing interests (related financial activities outside the submitted work). More information about this disclosure can be found on the JMLR website.}
The authors thank Juha Harviainen and Jack Kuipers for fruitful discussions on algorithmic ideas related to this work. The research was partially supported by the Research Council of Finland, grant 351156.}


\newpage

\appendix

\section{MCMC Frameworks Related to Duet Sampling}\label{sec:imcmc}

We place duet sampling within a broader class of algorithms for constructing reversible Markov chains. First, we show that duet sampling can be viewed as an instantiation of involutive MCMC (iMCMC) of \citet{Neklyudov2020}. Then we continue by observing that iMCMC itself is a special case of the more general auxiliary-variable Metropolis--Hastings algorithm of \citet{storvik2011}. 

\begin{table}[b!]
\begin{tabular}{lr}
\begin{minipage}{0.46\linewidth}
\begin{algorithm}[H]
\caption{iMCMC (single step)}
\begin{algorithmic}[1]
\State \textbf{input} target density $\pi(x)$ 
\State \textbf{input} proposal distribution $q(v|x)$ 
\State \textbf{input} involution $f(x, v)$ 
\State \textbf{input} current state $x$ 
\State sample $v \sim q(v|x)$
\State propose $(x',v') := f(x,v)$
\State compute the acceptance probability
\bes
	\alpha :=  \min\left\{1,\,
	\frac{\pi(x')\, q(v'|x')}{\pi(x)\, q(v|x)} \left| \frac{\partial f(x,v)}{\partial [x,v]} \right|
	\right\}
\ees
\State \textbf{output} %$x'$ with probability $\alpha$, and $x$ otherwise
\bes
	\begin{cases}
	x' & \text{with probability } \alpha\,, \\
	x  & \text{with probability } 1 - \alpha
	\end{cases}
\ees
\end{algorithmic}\label{alg:imcmc}
\end{algorithm}
\end{minipage}
\hspace{6pt} & \hspace{6pt}
\begin{minipage}{0.46\linewidth}
\begin{algorithm}[H]
\caption{Duet sampling (single step)}
\begin{algorithmic}[1]
\State \textbf{input} target density $\pi(x)$
\State \textbf{input} proposal distributions $\{q_k\}_{k \in L}$ 
\State \textbf{input} pairing $\{\beta_{kl}\}$ on $\{q_k\}$
\State \textbf{input} current state $x$ 
\State sample $(x', k,l) \sim \beta_{kl}(x) q_k(x'|x)$
\State \Comment{\emph{New state $x'$ got already proposed}}
\State compute the acceptance probability
\bes
\alpha:= \min\left\{1,\,
\frac{\pi(x')\, \beta_{lk}(x')\,q_l(x|x')}{\pi(x)\, \beta_{kl}(x)\,q_k(x'|x)}\right\}
\ees
\State \textbf{output} %$x'$ with probability $\alpha$, and $x$ otherwise
\bes
	\begin{cases}
	x' & \text{with probability } \alpha\,, \\
	x  & \text{with probability } 1 - \alpha
	\end{cases}
\ees
\end{algorithmic}\label{alg:ppmcmc}
\end{algorithm}
\end{minipage}
\end{tabular}
\end{table}

\subsection{Involutive MCMC}

\citet{Neklyudov2020} provide iMCMC as a unifying perspective encompassing a broad class of algorithms, 
from basic Metropolis--Hastings \citep{Hastings1970} to reversible-jump MCMC \citep{Green1995} and sample-adaptive MCMC \citep{zhu2019}, 
%including Metropolis--Hastings \citep{Hastings1970}, Hybrid Monte Carlo \citep{duane1987}, sample-adaptive MCMC \citep{zhu2019}, lifted MH \citep{turitsyn2011}, and reversible-jump MCMC \citep{Green1995}, 
within a common mathematical structure. It also offers a principled approach for building new samplers or extending existing ones. 

To construct a Markov chain with stationary density $\pi(x)$, iMCMC takes as input a user-defined involutive transformation $f(x, v)$, i.e., deterministic mapping that is its own inverse, defined on an extended state space augmented with an auxiliary variable $v$. Combined with a proposal distribution $q(v|x)$ and an appropriate acceptance rule, one obtaines a transition kernel that satisfies detailed balance. Algorithm~\ref{alg:imcmc} shows a single step of iMCMC.

To see how duet sampling is obtained as a special case of iMCMC, we write duet sampling in a fashion similar to iMCMC in Algorithm~\ref{alg:ppmcmc}. We choose the auxiliary variable $v$ and the involution $f(x, v)$ in a particular way: we put $v := (x', k, l)$ and $f(x, v) := (x', v')$, where $v' = (x, l,  k)$. Note that this involution simply swaps $x$ and $x'$, as well as $k$ and $l$, i.e., $f(x; x', k, l) = (x'; x, l, k)$; the Jacobian of $f$ equals $1$. 

\subsection{Auxiliary-Variable Metropolis--Hastings}

To place iMCMC in a broader context, we briefly review the generalized auxiliary-variable Metropolis--Hastings framework of \citet{storvik2011}. Here, Markov chains are constructed on an extended state space by augmenting the target distribution with auxiliary variables and defining proposals through arbitrary transformations on the augmented space. \citet{Neklyudov2020} mention this framework as earlier work related to iMCMC, but they do not review the relationship in detail.  

The acceptance probability is obtained by incorporating the forward and reverse transition probabilities induced by the proposed transformation, yielding a Metropolis--Hastings correction that guarantees detailed balance with respect to the augmented target distribution. The augmented target distribution is constructed to have the original target distribution as its marginal. The resulting Markov chain is invariant with respect to the augmented target, and therefore preserves the desired target distribution after integrating out the auxiliary variables. The acceptance probability within this framework can be written as

\be\label{eq:A1}  
	\alpha = \min\left\{1,\; \frac{\pi(x')\,h(v'| x') \,q(x,v|x',v')}{\pi(x)\,h(v | x)\,q(x',v'|x,v)}\right\}\, .
\ee
Here, $x$ denotes the current state, $v$ is the auxiliary variable; $\pi(x)$ is the target distribution of interest, and $h(v |x)$ is an auxiliary conditional density defining the augmented target~$\tilde{\pi}(x,v)=\pi(x)h(v| x)$. Importantly, as previously stated, $\tilde{\pi}(x,v)$ marginalizes to $\pi(x)$.

The iMCMC framework is a particular case of this construction, obtained by deterministically generating $(x',v')$ from $(x,v)$ via an involutive mapping $f$, and by setting $h(v| x)=q(v|x)$. We introduce infinitesimal volume elements in order to formulate a change of variable on $q(x,v|x',v')$, since transformations act on the underlying reference measure rather than directly on density functions. The change-of-variables formula then gives
\be\label{eq:A2}  
	q(x,v|x',v')\,\dd x\,\dd v=q(x',v'|x,v)\,\dd x'\,\dd v'\,,
\ee
with
\be\label{eq:A3}   
	\dd x'\,\dd v' = \left| \frac{\partial f(x,v)}{\partial [x,v]} \right|\,\dd x\,\dd v \,.
\ee
By substituting \eqref{eq:A3} into \eqref{eq:A2} and canceling out $\dd x\,\dd v$, we get 
\be\label{eq:A4}  
	q(x,v|x',v') = q(x',v'|x,v)\,\left|\frac{\partial f(x,v)}{\partial (x,v)} \right|\, .
\ee
By substituting \eqref{eq:A4} into \eqref{eq:A1} and setting $h(v |x)=q(v |x)$, the acceptance probability in \eqref{eq:A1} reduces to the iMCMC acceptance probability given in Algorithm \ref{alg:imcmc}.

\section{Normalizing Constants of the EC and AC Moves}\label{sec:norm}

We explicitly derive the normalizing constants of the proposal distribution $q_m(G' | G)$ of the edge-constrained(EC) and ancestor-constrained (AC) moves \eqref{eq:q1} and show that they are of the forms detailed in Section \ref{sec:pru}, thereby allowing faster approximate computation. Specifically, the requirement is that the terms can be expressed, for some sets $T,U\subseteq N$, as either an \emph{interval sum} with $|T|\leq 1$ or an \emph{intersection sum} of $\pi_i$, defined, respectively, as
\bes \label{eq:sums}
\sum_{T \subseteq J \subseteq U} \pi_i(J)
\qquad \text{and} \qquad
\sum_{J \subseteq U : J \cap T \neq \emptyset} \pi_i(J)\, .
\ees
It is immediate that, in order to preserve acyclicity, the descendants of a node cannot be considered as possible parents of that node. We therefore omit this condition when describing the individual cases. Recall that the set of descendants of node $j$, denoted by $D_j$, includes $j$ itself by definition. Furthermore, as stated in Section \ref{sec:ecpr}, the new parent set of node $i$ is sampled before that of node $j$.

\subsection{The EC Move}

We group the normalizing constants according to the constraint imposed. The add $ij$ move and reverse-to $ij$ moves impose the same constraints on the new DAG.

\newlength{\Uwidth}
\settowidth{\Uwidth}{$U = N \setminus D_j(G_i') = N \setminus D_j(G_0)$}

\subsubsection{Add $ij$ and Reverse-to $ij$}

%We require $ij\in G'$.
\begin{itemize}
\item Since the descendants of $j$ in $G_0$ cannot be parents of~$i$, the terms $Z_i^\text{add}(G_0)$ and $Z_i^\text{rev}(G_0)$ are interval sums with
\begin{flalign*}
\hspace*{1cm}
&\makebox[\Uwidth][l]{$U = N \setminus \big( D_i(G_0) \cup D_j(G_0) \big)$}
\hspace{1cm}
\text{and}
\hspace{1cm}
T = \emptyset \,.&
\end{flalign*}

\item Since $i$ must be a parent of $j$, the terms $Z_j^\text{add}(G_i')$ and $Z_j^\text{rev}(G_i')$ are interval sums with
\begin{flalign*}
\hspace*{1cm}
&\makebox[\Uwidth][l]{$U = N \setminus D_j(G_i')$}
\hspace{1cm}
\text{and}
\hspace{1cm}
T = \{ i \}\,. &
\end{flalign*}
\end{itemize}

\subsubsection{Remove $ij$}

%\daniele{We require $ij\notin G'$ and $ji \notin G'$.} 
\begin{itemize}
\item  Since $j$ cannot be a parent of $i$ in $G'$, the term $Z_i^\text{rem}(G_0)$ is an interval sum with
\begin{flalign*}
\hspace*{1cm}
&\makebox[\Uwidth][l]{$U = N \setminus \big( D_i(G_0) \cup \{j\} \big)$}
\hspace{1cm}
\text{and}
\hspace{1cm}
T = \emptyset \,. &
\end{flalign*}

\item Since $i$ cannot be a parent of $j$ in $G'$, the term $Z_j^\text{rem}(G_i')$ is an interval sum with
\begin{flalign*}
\hspace*{1cm}
&\makebox[\Uwidth][l]{$U = N \setminus \big( D_j(G_i') \cup \{i\} \big)$}
\hspace{1cm}
\text{and}
\hspace{1cm}
T = \emptyset \,.&
\end{flalign*}

\end{itemize}

\subsection{The AC Move}

Again, we group the normalizing constants according to the constraint imposed, even thought now the normalizing constants for node $i$ are equal for all moves. 

\subsubsection{Add $i\rightsquigarrow j$ and Reverse-to $i\rightsquigarrow j$}

\begin{itemize}

\item Since the descendants of $j$ in $G_0$ cannot be parents of $i$ in $G'$, the terms $Z_i^\text{add}(G_0)$ and $Z_i^\text{rev}(G_0)$ are interval sums with
\begin{flalign*}
\hspace*{1cm}
&\makebox[\Uwidth][l]{$U = N \setminus \big( D_i(G_0) \cup D_j(G_0) \big)$}
\hspace{1cm}
\text{and}
\hspace{1cm}
T = \emptyset \,. &
\end{flalign*}

\item Since the parent set of $j$ in $G'$ must intersect $D_i(G'_i)$, the terms $Z_j^\text{add}(G_i')$ and $Z_j^\text{rev}(G_i')$ are intersection sums with
\begin{flalign*}
\hspace*{1cm}
&\makebox[\Uwidth][l]{$U = N \setminus D_j(G_i') = N \setminus D_j(G_0)$}
\hspace{1cm}
\text{and}
\hspace{1cm}
T = D_i(G_i') = D_i(G_0) \,. &
\end{flalign*}

\end{itemize}

\subsubsection{Remove $i\rightsquigarrow j$}

\begin{itemize}

\item Since the descendants of $j$ in $G_0$ cannot be parents of $i$ in $G'$, the term $Z_i^\text{rem}(G_0)$ is an interval sums with
\begin{flalign*}
\hspace*{1cm}
&\makebox[\Uwidth][l]{$U = N \setminus \big( D_i(G_0) \cup D_j(G_0) \big)$}
\hspace{1cm}
\text{and}
\hspace{1cm}
T = \emptyset \,. &
\end{flalign*}

\item Since the descendants of $i$ in $G'_i$ cannot be parents of $j$ in $G'$, the term $Z_j^\text{rem}(G_i')$ is an interval sum with
\begin{flalign*}
\hspace*{1cm}
&\makebox[\Uwidth][l]{$U = N \setminus \big( D_j(G_i') \cup D_i(G_i') \big)$}
\hspace{1cm}
\text{and}
\hspace{1cm}
T = \emptyset \,. &
\end{flalign*}

\end{itemize}

\section{Convergence of the Chains Generated by the New Moves}\label{sec:apec}

We show that the Markov chains obtained by the edge-constrained (EC), ancestor-constrained (AC) and unconstrained sequential resampling (USR) moves converge to the stationary distribution  $\pi(G)$. Convergence is guaranteed when the chain is irreducible, aperiodic, and satisfies the detailed balance condition. Since the moves fall within the duet sampling framework, detailed balance is ensured by Theorem~\ref{thm:inv}. 

To establish irreducibility and aperiodicity,  we will rely on the following assumptions about the target distribution:
\begin{alignat*}{3}
\text{(modularity)}      && \quad \pi(G)
                         &\propto \prod_i \pi_i\big(P_i(G)\big)
                         && \quad \text{for all DAGs $G$}\,,\\
\text{(positive empty)}  && \pi(G^0)
                         &> 0
                         && \quad \text{for the empty DAG $G^0$}\,.
\end{alignat*}
In fact, we will only need the latter assumption for EC and AC. To avoid degenerate cases, we assume that there are (at least) two DAGs $G^1$ and $G^2$ with $\pi(G^1) > 0$ and $\pi(G^2) > 0$. 

Naturally, we will only consider the variants of EC, AC, and USR that are obtained using effective pairings (Propositions~\ref{prop:ec2}, \ref{prop:usr2}, \ref{prop:usr3}). In other words, we assume the following: 
\begin{alignat*}{3}
\text{(effectiveness)}   && \quad \beta_{kl}(G)q_k(G' | G)>0
                         &\implies \beta_{lk}(G')q_l(G | G')>0
                         && \quad \text{for all $G$, $G'$, $k$, and $l$}\,.
\end{alignat*}

\subsection{Irreducibility}\label{sec:ecirr}

A Markov chain is irreducible if every state $G^2$, with $\pi(G^2)>0$, can be reached from every other state $G^1$, with $\pi(G^1)>0$, in a finite number of steps with positive probability. 

\subsubsection{EC and AC}

Consider first the EC move. State $G^2$ can be reached from $G^1$ as follows: First, reach the empty graph $G^0$ by a finite number of remove $ij$ moves, each assigning the empty parent sets to nodes $i$ and $j$. Second, reach $G^2$ from $G^0$ by a finite number of add $ij$ moves, where $ij$ is selected such that edge $ij$ is absent in the intermediate graph but present in $G^2$, and assigning nodes $i$ and $j$ their parent sets in graph $G^2$. 

By the two assumptions, $\pi(G') > 0$ for every intermediate graph $G'$. Consequently, each intermediate graph is proposed with positive probability. Since the pairing is effective (Proposition~\ref{prop:ec2}), also the respective transition probability is positive. 

For the AC move, the proof is analogous, by analyzing the ancestor relations between nodes rather than the edges connecting them.

\subsubsection{USR}

To reach $G^2$ from $G^1$, instead of going via the empty graph, we directly replace the parent set of each node by its parent set in $G^2$. This is possible if processing the nodes in a topological ordering $\rho$ of $G^2$, such that for every edge $ij$ in $G^2$, node $i$ precedes $j$ in $\rho$. Supposing the USR move uses block size $s$, we partition $\rho$ into segments of size $s$; if necessary, the final segment may overlap with the preceding one to ensure that it also has size $s$. 

The key observation is that for each node $j$, after updating the parent sets for all nodes that precede $j$ in $\rho$, the parent set $P_j(G^2)$ is proposed with positive probability. This is because its local score is positive, by modularity, and because it cannot create a cycle, as in the obtained intermediate graph the ancestors of $j$ are the same as in $G^2$. Since the pairing is effective, the transition probability to each intermediate DAG is positive. 

\subsection{Aperiodicity}\label{sec:ecape}

A Markov chain is aperiodic if, for every state $G$ with $\pi(G)>0$, the greatest common divisor of the lengths of all possible return paths to $G$ equals 1. Aperiodicity could be easily ensured by modifying the chain so that it stays in the current with some small positive probability. However, below we prove aperiodicity without assuming such modifications, jointly for the EC and AC moves, and separately for the USR move.

\subsubsection{EC and AC}

Since the chain is irreducible, it suffices to show that there exists a DAG $G$ with $\pi(G) > 0$ such that the chain can return to $G$ with positive probability in both 2 and 3 steps.

Consider first the EC move. By the assumptions, we have $\pi(G^0) > 0$ and $\pi(G^1) > 0$ for the empty DAG $G^0$ and some other DAG $G^1$ in which only one node $j$ has a nonempty parent set; we may assume the parent set contains node $i$. Clearly, the transitions from $G^0$ to $G^1$ and back to $G^0$ have positive probabilities (by the add $ij$ and  remove $ij$ moves).

Now, suppose there is a third DAG $G^2$ with $\pi(G^2) > 0$ in which only node $i$ has a nonempty parent set that contains node $j$. Clearly the transition from $G^0$ to $G^2$ occurs with positive probability (by the add $ji$ move). But also the transition from $G^2$ to $G^1$ occurs with positive probability (by the reverse-to $ij$ move). Thus, the chain can return with positive probability to $G^0$ in exactly 3 steps.

For the AC move, the proof is analogous, since any DAG that is reached from $G$ via an EC move can also be reached from $G$ via an AC move.

\subsubsection{USR}

It suffices to observe that at any DAG $G$ with $\pi(G) > 0$, the same DAG $G$ is proposed and accepted as the next state with positive probability. This holds because no constraints other than acyclicity are imposed on the new parent sets.

\section{Comparison Between USR Move Variants}\label{sec:comp}

This section compares the properties of the involutive variants of the unconstrained sequential resampling (USR) move. We show that the use of the involution $f(\sigma)=\sigma^{\text{rev}}$ encourages the acceptance of high-scoring DAGs in which the ancestor relationships between some nodes in the current DAG are reversed, thereby enhancing the mixing of the chain. In contrast, the USR-id variant discourages such reversals, instead favoring the acceptance of DAGs in which ancestor relationships are maintained.

\subsection{Transition Probabilities of the USR-rev and USR-id Moves}

We will show the said difference of USR-rev and USR-id by comparing their transition probabilities from any current DAG $G$ to any proposed DAG $G'$. While we will assume for simplicity that the block size is $|S|=2$, the findings are seen to generalize also to $|S|>2$. In what follows, we let $S = \{i, j\}$. 

To simplify the analysis, we assume that $P'_i\neq P_i$, $P'_j\neq P_j$, and we set $\beta_\sigma=\beta_{f(\sigma)}=\beta$ for all $\sigma$. We adopt the notation of the EC and AC moves, for it is more concise and also applies to the USR move when $|S|=2$. Thus, we denote the intermediate DAGs obtained by incorporating the parent sets $P'_i$ and $P'_j$ into $G_0$ by  $G'_i$ and $G'_j$, respectively. Since we only compare variants that leverage involutive pairings, we denote the acceptance ratio \eqref{eq:a2} by $\alpha_{ij}^f$, where $ij$ is the node ordering used in the forward path to $G'$, and $f$ is the involution that specifies the node ordering used in the backward path to~$G$.

Recall that the USR move variants share the same proposal mechanism and differ only in their acceptance ratios. The transition probability from~$G$ to $G'$ is
\begin{equation}\label{eq:tp}
    p^f(G'|G)=\beta \,q_{ij}(G'|G) \, \alpha_{ij}^{f}+\beta \, q_{ji}(G'|G) \, \alpha_{ji}^{f} \,,
\end{equation}
where
\bes
	q_{ij}(G'|G) = \frac{\pi_i(P'_i)\pi_j(P'_j)}{Z_i(G_0)Z_j(G'_{i})}
	\quad \text{and} \quad
	q_{ji}(G'|G) = \frac{\pi_j(P'_j)\pi_i(P'_i)}{Z_j(G_0)Z_i(G'_{j})}\,.
\ees
With $f = \textrm{id}$, the acceptance probabilities are 
\be\label{eq:tpid}
	\alpha_{ij}^\text{id} = \min \bigg\{1,\frac{Z_j(G'_{i})}{Z_j(G_{i})}\bigg\}
		\quad \text{and} \quad
	\alpha_{ji}^\text{id} \min \bigg\{1,\frac{Z_i(G'_{j})}{Z_i(G_{j})}\bigg\}\,.
\ee
With $f = \textrm{rev}$, the acceptance probabilities are 
\be\label{eq:tprev}
	\alpha_{ij}^\text{rev} = \min \bigg\{1,\frac{Z_i(G_0)Z_j(G'_{i})}{Z_j(G_0)Z_i(G_{j})}\bigg\}
		\quad \text{and} \quad
	\alpha_{ji}^\text{rev} \min \bigg\{1,\frac{Z_j(G_0)Z_i(G'_{j})}{Z_i(G_0)Z_j(G_{i})}\bigg\}\,.
\ee

To compare $p^{\text{id}}(G'|G)$ to $p^{\text{rev}}(G'|G)$, we compare the normalizing terms $Z_i$ and $Z_j$ of the varying DAGs appearing in the corresponding acceptance ratios. We make use of the fact that $Z_i$ grows with the set of candidate parents of node $i$, i.e., the set of non-descendants of $i$ in the DAG at hand: If $D_i(G^1)\subseteq D_i(G^2)$, then $Z_i(G^1)\geq Z_i(G^2)$, for any $G^1$ and $G^2$. Also, if $D_i(G^1)= D_i(G^2)$, then $Z_i(G^1)=Z_i(G^2)$. Thus, instead of directly considering the normalizing constants, we may compare the sets of descendants of nodes $i$ and $j$ in the intermediate DAGs.

Note that if the new parent set of the first node in the permutation contains a descendant of the second node, then the descendant set of the second node is augmented with the descendant set of the first node. Conversely, if the new parent set of the first node contains no descendant of the second node, no such augmentation occurs. This distinction is the only factor that we take into account when we analyze the normalizing constants. Thus, to compare the transition probabilities of the two move variants, we need to first characterize the ancestor relationships between the nodes $i$ and $j$ in the proposed DAG, both in the forward and backward path to $G'$ and~$G$, respectively. 

To this end, for each scenario, we identify the descendant sets of $i$ and $j$ in the intermediate DAGs in terms of the sets $D_i(G_0)$, $D_j(G_0)$, or their union, as displayed in Table~\ref{tab:desc}. We omit from the table the descendant sets of $i$ and $j$ in the orphaned DAG $G_0$, which are, by definition, $D_i(G_0)$ and $D_j(G_0)$. To obtain the descendant sets of $i$ and $j$ in the intermediate DAGs along the backward transition from $G'$ to $G$, we simply replace $G'$, $D_i(G'_j)$, and~$D_j(G'_i)$ in Table~\ref{tab:desc} with $G$, $D_i(G_j)$, and~$D_j(G_i)$, respectively. 

The following proposition covers all the nine cases corresponding to the three possible ancestor configurations for both $G$ and $G'$. However, in the statement we omit the cases that follow by symmetry, that is, by swapping $i$ and $j$. 

\begin{table}[t!]
\centering
\begin{tabular}{l
                >{\centering\arraybackslash}m{3.2cm}
                >{\centering\arraybackslash}m{3.2cm}
                >{\centering\arraybackslash}m{4.5cm}}
\toprule
& \multicolumn{3}{c}{Ancestor relationship between $i$ and $j$ in $G'$} \\
\cmidrule(lr){2-4}
Descendants
&
$i \rightsquigarrow j$
&
$j \rightsquigarrow i$
&
$i \not\rightsquigarrow j$, $j \not\rightsquigarrow i$
\\
\midrule
$D_i(G'_j)$ & $D_i(G_0) \cup D_j(G_0)$  & $D_i(G_0)$ & $D_i(G_0)$ \\
$D_j(G'_i)$ & $D_j(G_0)$  & $D_i(G_0)\cup D_j(G_0)$ & $D_j(G_0)$ \\
\bottomrule
\end{tabular}
\caption{Descendants of nodes $i$ and $j$ in the intermediate DAGs $G'_i$ and $G'_j$ along the forward path from $G$ to $G'$.}
\label{tab:desc}
\end{table}

\begin{samepage}
\begin{proposition}
For all DAGs $G$ and $G'$ and nodes $i$ and $j$, we have 
\begin{description}[align=right, labelwidth=2em, leftmargin=!]
\item[(i)] $p^{\mathrm{rev}}(G'| G) = p^{\mathrm{id}}(G'| G)$ if $i\not \rightsquigarrow j$ and $j \not \rightsquigarrow i$ in $G$ or $G'$; 
\item[(ii)] $p^{\mathrm{rev}}(G'| G) \geq p^{\mathrm{id}}(G'| G)$ if $i\rightsquigarrow j$ in $G'$ and $j \rightsquigarrow i$ in $G$;
\item[(iii)] $p^{\mathrm{rev}}(G'|G) \leq p^{\mathrm{id}}(G'|G)$ if $i\rightsquigarrow j$ in both $G$ and $G'$.
\end{description}
\end{proposition}
\end{samepage}

\begin{proof}
We prove the claims separately.
\begin{description}[align=right, labelwidth=2em, leftmargin=!]
\item[(i)] Suppose first that $i \not\rightsquigarrow j$ and $j \not\rightsquigarrow i$ in $G$. For all the possible ancestor configurations between $i$ and $j$ in $G'$, by Table \ref{tab:desc}, we have $D_i(G_j)=D_i(G_0)\subseteq D_i(G'_j)$ and $D_j(G_i)=D_j(G_0)\subseteq D_j(G'_i)$. Thus, we obtain $\alpha_{ij}^\text{id}=\alpha_{ji}^\text{id}=\alpha_{ij}^\text{rev}=\alpha_{ji}^\text{rev} =1$, which implies  $p^{\mathrm{rev}}(G'| G) = p^{\mathrm{id}}(G'| G)$.

Suppose then that $i \not\rightsquigarrow j$ and $j \not\rightsquigarrow i$ in $G'$. For all the possible ancestor configurations between $i$ and $j$ in $G$, by Table \ref{tab:desc}, we have $D_i(G'_j)=D_i(G_0)$ and $D_j(G'_i)=D_j(G_0)$, yielding $\alpha_{ij}^\text{id}=\alpha_{ij}^\text{rev}$ and $\alpha_{ji}^\text{id}=\alpha_{ji}^\text{rev}$, respectively. Thus, $p^{\mathrm{rev}}(G'| G) = p^{\mathrm{id}}(G'| G)$.

\item[(ii)] Suppose that $i \rightsquigarrow j$ in $G'$ and $j \rightsquigarrow i$ in $G$. By Table \ref{tab:desc} we have $D_j(G'_i)=D_j(G_0)\subseteq D_j(G_i)$ and $D_i(G_j)=D_i(G_0)\subseteq D_i(G'_j)$, which imply $\alpha_{ij}^\text{id}=\alpha_{ij}^\text{rev}=1$ and $\alpha_{ji}^\text{id}\leq\alpha_{ji}^\text{rev}$. Consequently, $p^{\mathrm{rev}}(G'|G)\geq p^{\mathrm{id}}(G'|G)$.

\item[(iii)] Suppose that $i \rightsquigarrow j$ in both $G$ and $G'$. By Table \ref{tab:desc} we have $D_i(G_0)\subseteq D_i(G'_j)=D_i(G_j)$ and $D_j(G_0)= D_j(G'_i)=D_j(G_i)$, which imply $\alpha_{ij}^\text{rev}=\alpha_{ij}^\text{id}=1$ and $\alpha_{ji}^\text{rev}\leq\alpha_{ji}^\text{id}=1$. Consequently, $p^{\mathrm{id}}(G'|G)\geq p^{\mathrm{rev}}(G'|G)$.
\end{description}\vspace{-24pt}
\end{proof}

The above analysis also reveals distinct failure cases for both variants. Specifically, if $Z_i(G'_j) \ll Z_i(G_0)$, then $p^{\mathrm{id}}(G'| G) \approx 0$ under case~(ii), whereas $p^{\mathrm{rev}}(G'| G) \approx 0$ under case~(iii). To see this, observe that in either case,
\begin{equation}
\pi(P_i') < Z_i(G'_j) \ll Z_i(G_0),
\end{equation}
rendering $q_{ij}(G'| G) \approx 0$, thereby causing the first term in \eqref{eq:tp} to vanish. It remains to show that also the second term in \eqref{eq:tp} vanishes. For case (ii), since $Z_i(G_j) = Z_i(G_0)$, we have $\alpha_{ji}^{\mathrm{id}} = {Z_i(G_j')}/{Z_i(G_j)} \approx 0$. Similarly, for case (iii), since $Z_j(G_i) = Z_j(G_0)$, we have $\alpha_{ji}^{\mathrm{rev}} = {Z_i(G_j')}/{Z_i(G_0)} \approx 0$.

\subsection{Remarks and Generalization}\label{sec:rg}

The USR-id move encourages exploration of local modes that preserve the ancestor relationship between the nodes, while discouraging the acceptance of DAGs that would reverse ancestor relationships. Conversely, the USR-rev move encourages the reversal of ancestor relationships between nodes and consequently promotes more significant structural changes in the current DAG. This reduces autocorrelation and facilitates more efficient exploration of the posterior distribution. As discussed in Section~\ref{sec:pmt}, these parent-set resampling moves, by design, struggle to propose DAGs where the ancestor relationships are reversed. Therefore, variants that alleviate this limitation by promoting the acceptance of such reversals are preferred. 

An illustrative example can help highlight the advantageous properties of the USR-rev move compared to the USR-id move. Consider three high-scoring DAGs $G^1$, $G^2$, and $G^3$. The DAGs $G^1$ and $G^2$ share the same ancestor relation $i \rightsquigarrow j$, whereas $j \rightsquigarrow i$ in $G^3$. By performing the USR-id move, it is easy to transition from $G^1$ to $G^2$, but reaching $G^3$ from either $G^1$ or $G^2$ may be difficult. In contrast, the USR-rev move discourages the direct transition from $G^1$ to $G^2$, yet it still allows easy exploration of all three DAGs by favoring the transition from $G^1$ to $G^3$ and then to $G^2$.

These results naturally extend to the case $|S| > 2$. Namely, a similar analysis can be carried out by considering whether $i$ appears after $j$ or before $j$ in the permutation~$\sigma$. Again, one can verify that USR-rev encourages the acceptance of DAGs in which the ancestor relationships of the nodes in $S$ are reversed, thereby promoting significant structural changes.
 
In our experiments, we did not observe poor convergence due to the low acceptance rate of USR-rev that may arise when the same ancestor relationships between nodes in $S$ are preserved in $G'$. Should this behavior constitute a limitation, it could be mitigated by the USR-rnd variant, which allows sampling a backward permutation~$\tau$ that matches the reversed topological ordering of the current DAG $G$, thereby maximizing the transition probability.

While we have focused on the USR move, the results generalize also to other moves that are based on some strongly positive set of proposal distributions and that sample parent sets sequentially, e.g., the Markov blanket resampling (MBR) move. In Appendix \ref{sec:small} we empirically compare the performance of the rev, rnd, and id schemes for the USR and MBR moves on small networks.

\section{Computing Environment and Data Generation}\label{sec:env}
\label{se:environment}

To code of the new moves was written in \CC \xspace and compiled with \emph{GCC} using the flags \texttt{-std=c++17 -march=native -O3}. Each MCMC run was executed in a single-threaded configuration on a cluster with AMD EPYC 7452 processors (32 cores at 2 350 MHz) and 504 GB of RAM.

We selected benchmark Bayesian networks from Scutari's Bayesian Network Repository \citep{ScutariBNrepository}, distributed under the CC BY-SA 3.0 license, and generated our datasets by forward sampling 10\,000 data points from those.

The datasets and the source code are available at \url{https://github.com/sums-of-products/duetto}.

\section{Results on Small Networks}\label{sec:small}

In this section, we present the results obtained on two low dimensional datasets sampled from benchmark networks. We sampled 10\,000 data points from the {\sc{Asia}} and {\sc{Sachs}} benchmark networks, which are summarized in Table~\ref{tab:dataset_summary_small} in terms of the number of nodes, number of edges, and maximum in-degree (MID).
\begin{table*}[tb]
\centering
{\small
\setlength{\tabcolsep}{6.0pt}
\begin{tabular}{lccc}
\toprule
Network & \#nodes & \#edges & MID \\
\midrule
\BigCaps{Asia}   & 8   & 8   & 2 \\
\BigCaps{Sachs}  & 11  & 17  & 3 \\
\bottomrule
\end{tabular}
}
\caption{Summary of the benchmark networks used in the experiments.}
\label{tab:dataset_summary_small}
\end{table*}

The aim of the experiments is to show that the MCMC runs, obtained by performing the newly presented moves, asymptotically converge to the exact posterior distribution, which is accessible for small networks. Furthermore, both the Markov blanket resampling (MBR) and unconstrained sequential resampling (USR) moves sample new parent sets according to a given node ordering $\sigma$, and their respective sets of proposal distributions $\mathcal{Q}$ are strongly positive. We therefore compare the performance of their rev, rnd, and id variants, to assess whether the id variant is indeed inferior, as theoretically illustrated in Appendix \ref{sec:comp}.

The exact posterior distributions have been computed using \emph{Bayesian intervention distribution adjustment} (BIDA) \citep{Pensar2020} and \emph{modular DAG sampling} (MDS) \citep{Talvitie2019uai} algorithms. BIDA algorithm, which relies on the IDA method, takes $O(3^nn)$ to compute the posterior scores of all parent sets. However, in rare cases, BIDA might fail due to catastrophic cancellations. We encountered this problematic while computing the posterior distribution for the Sachs network. Hence, we relied on MDS in this specific scenario. MDS directly samples DAGs from the exact posterior. The theoretical basis lies in the recurrences of  \cite{Tian2009}, while the root-layering method by \cite{Kuipers2013, Kuipers2017} is used to sample DAGs. The whole process takes $O(3^n)$ for pre-processing and $O(2^n)$ for sampling. To generate the reference distribution, $10^6$ DAGs have been sampled.

For each algorithm, we performed ten MCMC runs, each consisting of $10^7$ sampled DAGs. No burn-in or thinning was applied. We used a sparse prior together with the BDeu scoring system. We introduced both notions in Section \ref{sec:exp}. Similarly to the experiments on large networks,  we set the prior probability of parent sets whose size exceeds the MID of the data-generating network to zero, and we employed the score pruning technique detailed in Section \ref{sec:pru} with $\epsilon=2^{-10}$. 

\begin{figure*}[t!]
\begin{center}
\setlength{\tabcolsep}{3pt}  % horizontal padding
\small  % smaller font size

\begin{tabular}{>{\raggedright\arraybackslash}m{0.09\textwidth}  % algorithm names
                >{\centering\arraybackslash}m{0.43\textwidth}  % Asia
                >{\centering\arraybackslash}m{0.43\textwidth}} % Sachs

% -------------------- Column headers --------------------
& \large \asia{} ($n = 8$) & \large \sachs{} ($n = 11$)   \\
\midrule
\addlinespace[12pt]

% \multicolumn{3}{c}{\Large \datasets{}} \\
% \midrule
%  & \large  & \large \sachs{} ($n = 11$) \\
% \addlinespace[24pt]

% ==================== Row 1 ====================
USR &
\includegraphics[width=\linewidth,height=0.21\textheight,keepaspectratio]{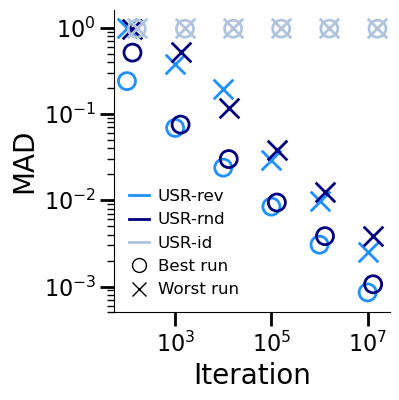} &
\includegraphics[width=\linewidth,height=0.21\textheight,keepaspectratio]{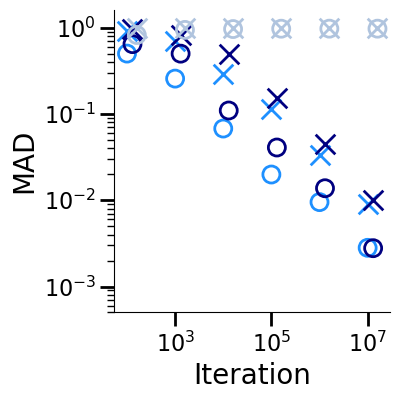} \\
\addlinespace[12pt]

% ==================== Row 2 ====================
EC \& AC &
\includegraphics[width=\linewidth,height=0.21\textheight,keepaspectratio]{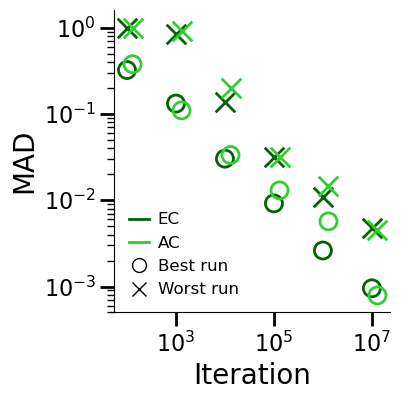} &
\includegraphics[width=\linewidth,height=0.23\textheight,keepaspectratio]{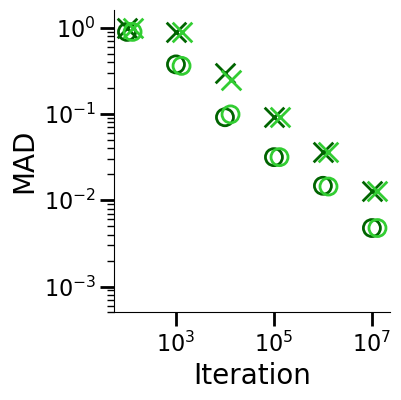} \\
\addlinespace[12pt]

% ==================== Row 3 ====================
MBR &
\includegraphics[width=\linewidth,height=0.21\textheight,keepaspectratio]{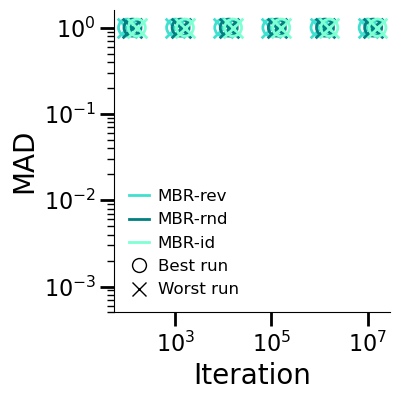} &
\includegraphics[width=\linewidth,height=0.21\textheight,keepaspectratio]{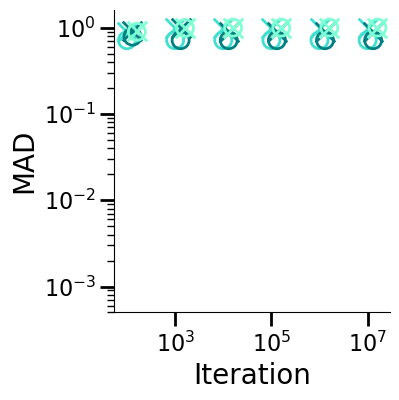} \\

\end{tabular}

\normalsize
\caption{The MAD of the best and worst runs of each algorithm on low-dimensional datasets sampled from benchmark networks. Columns correspond to datasets, while rows correspond to different algorithms.}
\label{fig:small}
\end{center}
\end{figure*}

To compare the performance of the algorithms, we report the \emph{maximum absolute deviation} (MAD), which is $|\hat{p}_{ij}-p_{ij}|$ over all edge probability estimates $\hat{p}_{ij}$ of a given MCMC run. Here $p_{ij}$ is the exact edge probability. In Figure \ref{fig:small}, we report the MAD of the best run and the worst run for each algorithm, which are the ones associated with larger and lower MAD respectively, computed after $10^7$ iterations. We better illustrate the asymptotic convergence of the new moves by using the logarithmic scale for both the MAD and iteration values.

Figure \ref{fig:small} shows that the edge-constrained (EC), the ancestor-constrained (AC), the USR-rev, and the USR-rnd moves correctly converge the exact posterior distribution, with the USR-rev move slightly outperforming the others. The id scheme performs poorly for both USR and MBR moves, with the corresponding MAD value being close to 1 also for the best runs in all possible scenarios. The MBR move fails to exhaustively approximate the posterior distribution on both networks; however, the MBR-rev and MBR-rnd moves still perform better than the MBR-id move on the Sachs network.

% \begin{figure*}[t!]
% \begin{center}
% \resizebox{\textwidth}{!}{
% \begin{tabular}{>{\centering\arraybackslash}m{0.45\textwidth}
%                 >{\centering\arraybackslash}m{0.45\textwidth}}

% % -------------------- Column headers --------------------
% \multicolumn{2}{c}{\Large \datasets{}} \\
% \midrule
% \Large \asia{} ($n = 8$) & \Large \sachs{} ($n = 11$) \\
% \addlinespace[16pt]

% % ==================== Row 1 ====================
% \multicolumn{2}{c}{\textit{BMH-rev, BMH-rnd, \& BMH-id}} \\
% \includegraphics[height=0.26\textwidth]{fig/asiabmh.png} &
% \includegraphics[height=0.26\textwidth]{fig/sachsbmh.png} \\
% \addlinespace[16pt]

% % ==================== Row 2 ====================
% \multicolumn{2}{c}{\textit{EC \& AC}} \\
% \includegraphics[height=0.26\textwidth]{fig/asiaec.png} &
% \includegraphics[height=0.26\textwidth]{fig/sachsec.png} \\
% \addlinespace[16pt]

% % ==================== Row 3 ====================
% \multicolumn{2}{c}{\textit{MBR-rev, MBR-rnd, \& MBR-id}} \\
% \includegraphics[height=0.26\textwidth]{fig/asiambr.png} &
% \includegraphics[height=0.26\textwidth]{fig/sachsmbr.png} \\

% \end{tabular}}
% \caption{
% \label{fig:small}
% MAD value of the best and worst runs of each algorithm, on low-dimensional datasets sampled from benchmark networks. Columns correspond to datasets, while rows correspond to different classes of samplers, grouped by algorithmic design.
% }
% \end{center}
% \end{figure*}

% \section{Additional simulation results}
% Table~\ref{tab:supp-data} lists additional simulation results; see also \citet{einstein} for a comparison. 

% \begin{table}[!h]
%     \centering
%     \caption{An Interesting Table.} \label{tab:supp-data}
%     \begin{tabular}{rl}
%         \toprule % from booktabs package
%         \bfseries Dataset & \bfseries Result\\
%         \midrule % from booktabs package
%         Data1 & 0.12345\\
%         Data2 & 0.67890\\
%         Data3 & 0.54321\\
%         Data4 & 0.09876\\
%         \bottomrule % from booktabs package
%     \end{tabular}
% \end{table}

% \section{Math font exposition}
% % NOTE: necessary when ptmx or no mathfont class option is given
% \providecommand{\upGamma}{\Gamma}
% \providecommand{\uppi}{\pi}
% How math looks in equations is important:
% \begin{equation*}
%     F_{\alpha,\beta}^\eta(z) = \upGamma(\tfrac{3}{2}) \prod_{\ell=1}^\infty\eta \frac{z^\ell}{\ell} + \frac{1}{2\uppi}\int_{-\infty}^z\alpha \sum_{k=1}^\infty x^{\beta k}\mathrm{d}x.
% \end{equation*}
% However, one should not ignore how well math mixes with text:
% The frobble function \(f\) transforms zabbies \(z\) into yannies \(y\).
% It is a polynomial \(f(z)=\alpha z + \beta z^2\), where \(-n<\alpha<\beta/n\leq\gamma\), with \(\gamma\) a positive real number.

\vskip 0.2in
\bibliography{paper}

\end{document}